\documentclass[11pt,letterpaper]{article}
\usepackage[a4paper, total={7in, 10in}]{geometry}

\usepackage[utf8]{inputenc}
\usepackage[T1]{fontenc}
\usepackage[english]{babel}
\usepackage{graphicx}
\usepackage{placeins}
\usepackage[font=footnotesize]{caption}
\usepackage{helvet}
\usepackage{authblk}
\usepackage[dvipsnames]{xcolor}
\usepackage{svg}
\usepackage{amsmath} 
\usepackage{amssymb} 
\usepackage{orcidlink} 
\usepackage[super,comma,sort&compress]{natbib}
\hypersetup{
    colorlinks=true,
    linkcolor=CornflowerBlue,
    citecolor=CornflowerBlue,
    urlcolor=CornflowerBlue,
    filecolor=CornflowerBlue,
    pdfborder={0 0 0}
}
\usepackage{cleveref}
\usepackage{enumitem}

\makeatletter
\renewcommand{\maketitle}{\bgroup\setlength{\parindent}{0pt}
\begin{center}
  \Large{\textbf{\@title}}
\end{center}
\begin{flushleft}
  \@author
\end{flushleft}\egroup}
\makeatother

\title{On Growth and Form, and Function: Reusable Regulatory Handles Control Phenotypic Variation}
\date{}

\author[1,$*$\orcidlink{0000-0001-7787-4839}]{Benedikt Hartl}
\author[2,\orcidlink{0000-0002-7440-3510}]{Milton L. Montero}
\author[2,\orcidlink{0009-0003-5765-9236}]{Marcello Barylli}
\author[2,3,\orcidlink{0000-0003-3607-8400}]{Sebastian Risi}
\author[1,4,$*$,\orcidlink{0000-0001-7292-8084}]{Michael Levin}

\affil[1]{Allen Discovery Center at Tufts University, Medford, MA, USA}
\affil[2]{IT University of Copenhagen, Denmark}
\affil[3]{Sakana AI, Japan}
\affil[4]{Wyss Institute for Biologically Inspired Engineering at Harvard University, Boston, MA, USA}

\affil[$*$]{Corresponding Authors: \href{mailto:hartl.bene.research@gmail.com}{hartl.bene.research@gmail.com} and \href{mailto:michael.levin@tufts.edu}{michael.levin@tufts.edu}}

\begin{document}
%%%%%%%%%%%%%%%%%%%%%%%%%%%%%%%%%%%%%%%%%%%
%%% FRONT MATTER %%%%%%%%%%%%%%%%%%%%%%%%%%
\maketitle


\textbf{Running Title:}
LoRA-NCA: low-rank regulatory control of phenotypic variation\\

%%%%%%%%%%%%%%%%%%%%%%%%%%%%%%%%%%%%%%%%%%%%%%%%%%%%%%%%%%%%%%%%%%%%%%%%%%%
%%% Abstract %%%%%%%%%%%%%%%%%%%%%%%%%%%%%%%%%%%%%%%%%%%%%%%%%%%%%%%%%%%%%%
%%%%%%%%%%%%%%%%%%%%%%%%%%%%%%%%%%%%%%%%%%%%%%%%%%%%%%%%%%%%%%%%%%%%%%%%%%%
\begin{abstract}
How phenotypic transformations are implemented by changes in the underlying regulatory dynamics remains a central question in developmental biology. Inspired by D’Arcy Thompson’s 1917 ``On Growth and Form'', we investigate whether coherent, large-scale geometric transformations of morphology can be encoded as low-dimensional modulations of a self-organizing developmental system. We use neural cellular automata (NCAs) as bio-inspired \textit{in silico} models of distributed development, in which a shared local regulatory network drives cellular development from a single seed toward a target morphology. To model how systematic modulations of cellular regulatory networks unfold into targeted changes in developmental outcomes, we apply low-rank adaptation (LoRA) to the regulatory parameters of pretrained NCAs. LoRA represents each adapted developmental program as a low-rank weight modulation of a fixed regulatory scaffold shared across adaptations. We show that horizontal and vertical scaling transformations of a fully grown 2D emoji phenotype can each be implemented by rank-one adaptations of the NCA’s regulatory weights. Linear combinations of these adaptations parametrically control the size of the resulting phenotype, generalize to scaling factors outside the training distribution, and compose with target-specific adapters. Strikingly, adaptations learned for a single phenotype transfer zero-shot across structurally and semantically diverse phenotypes trained relative to the same reference frame, while largely preserving their internal features. These results suggest that the learned adaptations represent reusable, system-level hyper-directions of scale rather than transformations tied to a particular morphology. From a static dataset of approximately 25,000 independently trained phenotype-specific NCA adapters with shared scaffold, we empirically discover latent low-dimensional hyper-directions that functionally address phenotypic variation such as scaling, style, or symmetrical fission of developmental outcomes. Together, our results provide a computational realization of D’Arcy Thompson's remarkable grid transformations in a 2D NCA---a minimal cybernetic tissue in which variations in shape and other traits of fully grown emoji phenotypes can be encoded, combined, and controlled through interventions along low-dimensional directions in regulatory weight space.
\end{abstract}

\textbf{Highlights:}
\newline
\begin{itemize}[nosep]
    \item LoRA is effective to fine-tune complex dynamical systems such as NCAs.
    \item LoRA can map phenotypic variation to regulatory modulation in cybernetic systems.
    \item The corresponding regulatory weight space has semantic geometry with functional hyper-directions.
    \item We train and empirically find universal hyper-directions in regulatory space, that generalize across different phenotypes with shared regulatory logic.
    \item LoRA decomposition provides a principled way to identify such low-rank regulatory handles for top-down control of cybernetic tissue, which has strong implications for evolvability and biological plasticity in general.
    \item D'Arcy Thompson's grid transformations (1917) can be controlled through such latent regulatory handles anticipated by Howard Pattee (1973).
\end{itemize}

%%% FRONT MATTER %%%%%%%%%%%%%%%%%%%%%%%%%%
%%%%%%%%%%%%%%%%%%%%%%%%%%%%%%%%%%%%%%%%%%%

%%%%%%%%%%%%%%%%%%%%%%%%%%%%%%%%%%%%%%%%%%%
%%% CONTENT %%%%%%%%%%%%%%%%%%%%%%%%%%%%%%%
\section{Introduction}
\begin{figure}[h]
    \centering
    \includegraphics[width=\linewidth]{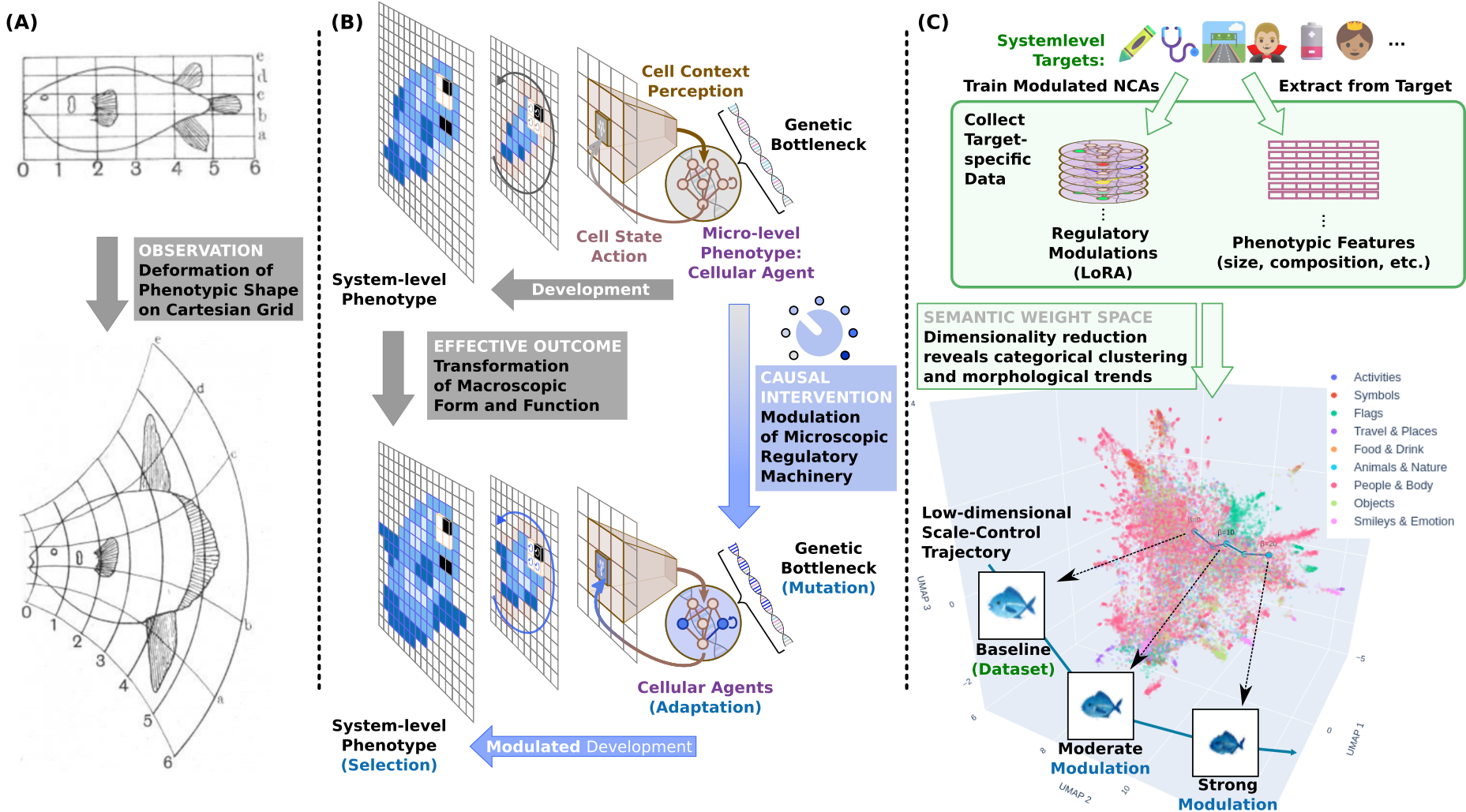}
    \caption{
    \textbf{Modulations of self-orchestrated morphogenesis canalize phenotypic transformations and construct a semantic weight space of the regulatory machinery.}
    (A) D'Arcy Thompson's Cartesian grid transformations relate phenotypic shapes, here using the example of a porcupinefish being transformed into a sunfish, reproduced from Ref.\citenum{thompson1917growth}; public domain.
    (B) Schematic flow diagram of how modulations at the regulatory level of self-orchestrated morphogenesis canalize phenotypic transformations. Multicellular morphogenesis can be modeled by neural cellular automata (NCAs) which comprise cellular agents whose numerical states are updated by a shared artificial neural network (ANN) that guides cellular perception and cell state updates within the local context of a multicellular grid. While cell states in NCAs resemble physiological states of biological cells, their ANNs operationalize gene-regulatory networks (GRNs) responsible for local intercellular decision-making to achieve collective system-level outcomes. Low-dimensional modulations of the regulatory parameters---ANN parameters, in our case---induce adapted cellular behavior at the microlevel, which can be canalized into effective system-level transformations.
    (C) Training NCAs on diverse emoji targets reveals a semantically structured and controllable weight space. In this work, we fine-tuned $\approx 25{,}000$ shared-scaffold NCAs via low-rank adaptation (LoRA). The LoRA weights organize into a semantic weight space, as visualized by the UMAP dimensional-reduction scatter plot that is colored by emoji group (see legend).
    Dominant principal components in the LoRA weight space control global target features such as phenotypic size; linearly translating the ANN of a baseline NCA parameters along those principal directions yields progressive high-level shape transformations while retaining phenotype-specific features.
    }
    \label{fig:introduction}
\end{figure}

In his 1917 book \textit{On Growth and Form}~\cite{thompson1917growth}, D'Arcy Thompson argued for systematically applying mathematical and physical principles to the biological world. He described the effects of scale on the shape of animals and plants, pointing out correlations between biological forms and mechanical phenomena.
Most famously, in the chapter ``The Comparison of Related Forms'', Thompson showed that morphological variations among related species can be understood through simple mathematical transformations.
By drawing a Cartesian grid around the anatomy of one species and deforming it, Thompson often obtained the characteristic shapes of other, related species such as certain kinds of fish, see \Cref{fig:introduction}~A.
The procedure goes as follows:
(1) Choose the morphology of a particular species.
(2) Draw a Cartesian grid around the morphology.
(3) Apply deformations to the grid via coordinate transformations.
(4) Carry along the corresponding transformation to the anatomical landmarks by the same map.
(Possibly 5) end up with the morphology of a different species.

D'Arcy Thompson pioneered the idea that biological organization is strongly governed by mathematical and mechanical principles.
These insights inspired great thinkers across biology, mathematics and the physical sciences, including Waddington, Turing and Kauffman, and became important precursors of modern evolutionary developmental biology (evo-devo).\cite{abzhanov2017Thompson}
Yet Thompson's transformations still lacked a causal developmental explanation, as emphasized by Arthur:\cite{Arthur2006Thompson} they neither identify which microscopic processes produce these anatomical differences nor where---and when---they must change during growth.
And despite major scientific advances in our understanding of growth and form in biological organization,
Thompson left behind a century-old inverse problem:\cite{Briscoe2017} if two anatomies are related by a simple grid transformation, what is the corresponding modulation of the developmental dynamics that physically generate them?
Put simply, what does the grid, and its global deformation, map onto in developmental and evolutionary biology?

Here, we use neural cellular automata (NCAs) as minimal models of cybernetic developmental material:\cite{mordvintsev_growing_2020, Hartl2025NCAs} artificial cells grow target morphologies from a single seed through repeated local communication and computation.
By applying Low-Rank Adaptation\cite{Edward2022LoRA} (LoRA) to their shared regulatory machinery, we test whether compact, yet distributed modulations of how the cells communicate and compute can redirect collective development toward coherent anatomical transformations (\Cref{fig:introduction}~B).
We demonstrate that Thompson's grid transformations can be represented and controlled by reusable low-dimensional directions in the parameter space of NCAs.
While LoRA-NCA serves as a computational framework for identifying such directions, our results suggest that analogous handles in the regulatory networks of biological cells may provide functional control over organism growth and form.

% History
Huxley showed how differential growth ratios of different body parts change the proportions of the entire organism, providing a developmental interpretation of Thompson's transformations.\cite{HUXLEY1924}
Waddington introduced canalization and later represented it through an ``epigenetic landscape'', describing how developmental trajectories converge toward stable phenotypic outcomes despite genetic and environmental perturbations.\cite{waddington1942canalization, waddington1957strategy}
Turing demonstrated how local reaction--diffusion dynamics generate spatial pattern.\cite{turing1952chemical}
Rosenblueth, Wiener and Bigelow described purposeful biological behavior through feedback, while Ashby showed how error correction and homeostatic adaptation sustain organization under perturbation;
together, these works established feedback-driven self-regulation as a ubiquitous principle of biological and cybernetic organization.\cite{rosenblueth1943behavior,Ashby1952}
Simon described complex systems as hierarchies of systems nested within systems---now called multiscale organization---characterized by nearly decomposable sub-systems.\cite{simon1962complexity}
Kauffman's random Boolean networks---early models of gene regulatory networks (GRNs)---showed how distributed regulatory dynamics can generate stable attractors.\cite{kauffman1969randomnetworks}
Pattee argued that control hierarchies achieve effective phenotypic control through a selective loss of detail, anticipating handles through which system-level outcomes can be redirected without micromanaging the detailed cellular or molecular dynamics.\cite{pattee1973control}
Conrad later showed that component redundancy---or degeneracy---allows phenotypes to remain stable while their underlying organization varies, creating room for evolutionary change.\cite{conrad1990evogeom}
Together, these ideas established biological form as a robust yet plastic outcome of self-regulated developmental dynamics operating across physical scales and evolutionary times.

% SoA
Bookstein later developed a quantitative method for studying biological shape and growth,\cite{Bookstein1978Thompson} while Stone showed that empirical morphospace analysis closely resembles Thompson's coordinate-transformation approach.\cite{Stone1997Thompson}
The field of geometric morphometrics now compares biological forms by registering homologous anatomical landmarks and decomposing the resulting deformation into principal warps.\cite{bookstein1989warps}
Evo-devo relates morphological variation to changes in development: heterochrony concerns changes in developmental timing and rate, while developmental modularity and evolvability describe how regulatory changes can redeploy conserved processes to generate viable phenotypic variation\cite{alberch1979size,Wagner1996,Gerhart2007}---evolution is tinkering.\cite{Jacob1977}
Mechanistic models connect particular morphologies to measured growth and mechanical parameters: differential edge growth explains for instance the opening of a lily,\cite{Liang2011} while relative growth and elasticity of the gut tube and mesentery predict intestinal looping across vertebrate species.\cite{Savin2011}
At the tissue scale, morphogenesis emerges from genetically regulated cellular growth and from force generation and transmission through cytoskeletal and adhesive machinery.\cite{thompson2021Thompson, Lecuit2011}
Mechanistic models thus define a forward map from developmental parameters and policies to morphology; identifying such parameters or policies for a prescribed morphology poses the corresponding inverse problem.
Within specified physical or computational models, inverse methods can compute anisotropic growth patterns that generate prescribed shapes in elastic bilayers\cite{vanRees2017} and infer local interaction rules and genetic networks for prescribed cell-cluster outcomes.\cite{deshpande2025engineering}
However, homomorphy---the conservation of anatomical form despite changes in its underlying developmental mechanism---means that form alone does not identify a unique regulatory implementation (both evolutionary and developmental processes can exhibit canalization to discrete anatomical attractors from distinct starting states or through different cell behavior mechanisms).\cite{Newman2019}
It remains unknown whether coherent transformations between related anatomies correspond to compact, reusable directions in the distributed regulatory dynamics of the governing multicellular substrate of life, and how such directions could be inferred from anatomical outcomes.

% Multiscale Competency Architecture
We now know that biology comprises layers within layers of organization, where the components of one layer form the substrate of the next\cite{simon1962complexity}.
It is increasingly recognized that the components of every layer have problem-solving competencies within their respective domains:
Even cells and cellular collectives solve problems, but in spaces less familiar than the 3D world of animal behavior:
individual cells regulate physiological, metabolic, and transcriptional states to achieve adaptive outcomes under a range of novel and unpredictable stressors, while cellular collectives navigate morphospace---the space of anatomical possibilities---despite injury and challenges that require remodeling or novel transition states to achieve the same target morphology.\cite{Friston2015KnowingOnesPlace,fields2020scale,fields2022competency,Levin2023DAM,Hartl2026RemappingNavigation}
When a planarian flatworm, or an early mammalian embryo, is cut in half, for example, each fragment reliably regenerates the missing structures to form a complete body, despite starting from an anatomical configuration that does not occur during normal development.\cite{reddien2004fundamentals,levin2019planarian}
Likewise, the scrambled craniofacial organs of ``Picasso tadpoles'' rearrange during metamorphosis as their tissues follow novel paths from inappropriate starting positions toward the correct large-scale configuration, yielding largely normal frog faces and thus demonstrating problem-solving at the level of entire organs.\cite{Vandenberg2012}
Crucially, the hardware and software of the biological substrate are intrinsically unreliable; in the presence of molecular noise, mutations, and changing environmental conditions, genetic information cannot simply be treated by the cells as prescribing fixed sequences of microscopic events, but must be interpreted by cells in their current physiological, anatomical and environmental context.\cite{Levin2023DAM}
Gene regulatory networks regulate intracellular states over developmental time,\cite{Davidson2002} while physiological, biomechanical, and bioelectric networks mediate communication across cells and tissues that coordinate long-distance events toward a coherent species-specific endpoint on multiple scales.\cite{levin2018bioelectriccode, whited2019bioelectrical}
The genome therefore does not encode morphology as a static blueprint but instantiates a generative model of the organism, whose distributed regulatory structure is decoded through development.\cite{mitchell2024genomic, hartl2025generativegenome}
Cells are not passive building blocks but agents whose regulation and communication form a distributed problem-solving layer between the genotypic regulatory machinery and phenotypic form and function---biology is a \textit{multiscale competency architecture}, implemented by an agential cellular substrate.\cite{Levin2023DAM, Levin2026MachinesUpCognitionDown}
Controlling anatomy becomes a question of finding compact intervention handles in these multiscale regulatory processes that direct the collective toward different morphological outcomes. Thus, it is important to develop models of morphogenesis and evolutionary shape change which include not merely a complex genotype-phenotype mapping, but more bio-realistic components of minimal decision-making agents at multiple scales of organization, whose presence requires active alignment toward large-scale setpoints and offers novel modes of control at the system level. 

NCAs\cite{mordvintsev_growing_2020} can be effectively used as minimal models of the distributed organization of agential biological tissue,\cite{Hartl2025NCAs} providing the biologist with a well-studied framework in which local agents (e.g., cells) obey rules (enabling the needed emergence and complexity) but, unlike conventional cellular automata models of developmental patterning,\cite{young1983CASkinPattern, cocho1987CAColorPattern, ermentrout1993CABio, silvia2003CA} their learnable update rules allow them to respond and change flexibly to continuously changing environmental contexts during their lifetime. This is bio-realistic, as the behavior of biological cells is dynamically regulated through gene-regulatory and biophysical networks within them.

NCAs comprise a grid of artificial cells, each maintaining a numerical state and an internal artificial neural network (ANN) with shared parameters across all cells.
These artificial cell states broadly represent physiological states of biological cells, while the ANN substitutes their regulatory machinery.
Across discrete developmental steps, each cell of the NCA perceives the state information of its Moore neighborhood to compute an update of its own state.
Through repeated interactions, the cells can collectively grow and maintain a target morphology from a single seed.
In their standard implementation,\cite{mordvintsev_growing_2020} one set of ANN parameters is trained for one target morphology (representing evolutionary tuning of control networks in cells that reliably achieve adaptive form and function at the organism scale).
Established developmental outcomes of NCAs can be ``hijacked'' through adversarial state-space attacks\cite{Cavuoti2022, Randazzo2021}, goal-guided NCAs supply cells with target encodings during growth,\cite{sudhakaran2022goalguided} EngramNCA stores and propagates target-specific information in private cell states,\cite{Guichard2025EngramNCA} and learned developmental pre-patterns can reliably condition a shared NCA toward different target morphologies.\cite{Montero2026MorphogenNCA}
In all of these cases, different outcomes are specified by changing the state information supplied to the cells while keeping a fixed set of trained regulatory parameters.
Here, we are interested in the possible semantic relations in the weight-space of NCAs.
To the best of our knowledge, only two prior contributions explicitly construct latent spaces of regulatory encodings for NCAs, either mapping image embeddings to target-dependent NCA parameter sets,\cite{Hernandez2021Manifolds} or by meta-learning an evolvable genotype-to-phenotype map.\cite{montero2024meta}
By contrast, we retain a pretrained NCA's regulatory program and ask whether minimal low-rank displacements of its parameters yield functionally persistent control over downstream developmental outcomes across morphologies---i.e., whether the remarkable coordinated shape change studied by Thompson can be represented by transferable low-dimensional directions in regulatory weight space (\Cref{fig:introduction}~A,B).
This would provide a mechanistic and computational account of what is happening when an organism’s ``grid'' is being deformed---by evolution, and someday soon, by bioengineers.

We hypothesize that coherent phenotypic variation is controlled by distributed but low-dimensional regulatory modes: coordinated changes across many parameters of the shared regulatory network that span only a small number of effective directions (\Cref{fig:introduction}~B).
To systematically map desired system-level variations to modulations at the regulatory-level, we use LoRA,\cite{Edward2022LoRA} keeping the baseline ANN parameters of an NCA frozen while trainable LoRAs bias the cellular state-update rule and thereby its developmental pathways toward target transformations at the phenotypic outcome.
We find that rank-one adaptations suffice to control horizontal and vertical scaling.
Linear modulation and combination of these adaptations generate non-linear, proportion- and feature-preserving phenotypic transformations, generalize to scaling factors outside the training distribution, and transfer zero-shot across different target phenotypes that share the same regulatory ANN scaffold.
We further train $\approx25{,}000$ phenotype-specific LoRA-NCAs with a shared scaffold and find dominant directions in their regulatory weight space associated with phenotypic extent, visual style, and symmetrical fission (cf. \Cref{fig:introduction}~C).

D'Arcy Thompson's scaling transformations are therefore represented as---and can be controlled by---low-dimensional latent variables in the regulatory machinery of the cybernetic tissue across a plethora of NCAs.
Such low-dimensional modulations can change macroscopic features without prescribing detailed cellular trajectories, providing a concrete realization of the effective multiscale control handles anticipated by Howard Pattee.\cite{pattee1973control}
Biologically, this suggests how evolution could access coherent transformations of organismal form and function through compact changes in the regulatory rules governing the cellular behavior and interactions.

\section{Methods}
\subsection{Neural Cellular Automata}
\label{sec:methods:nca}
\paragraph{Cellular Automata (CAs)} are dynamical systems implemented by a set of $i=1,\dotsc,N$ cells that typically interact in a local neighborhood $\mathcal{N}_i$, e.g., on a 1D or 2D spatial grid, to maintain cell-specific numerical state information of typically a discrete state set ${\boldsymbol x}_i^t\in\{c_1,\dotsc,c_\gamma\}$ over discrete time steps, $t\mapsto t+1$, to yield state updates, $\boldsymbol x_{i}^{t}\mapsto \boldsymbol x_i^{t+1}$, of the form $\boldsymbol x_{i}^{t+1}=u(\boldsymbol X_i^t)$;
the typically fixed (hardcoded) update function $u(\cdot)$ integrates the instantaneous state information $\boldsymbol X_i^t=\{\boldsymbol x_{i_\nu}^t\}_{{i_\nu}\in\mathcal{N}_i}$ of all cells $i_\nu$ within the local neighborhood $\mathcal{N}_i$ of cell $i$.\cite{vonNeumann1966, Wolfram2002}

\paragraph{Neural Cellular Automata (NCAs)} extend CAs by replacing the hardcoded update function $u(\cdot)\mapsto f_\theta(\cdot)$ with a more flexible artificial neural network (ANN), $f_\theta(\cdot)$, yielding a residual trainable update rule
\begin{equation}
    x_{i}^{t+1}=x_i^{t} + f_\theta(\boldsymbol X_i^t)
    \label{eq:nca}
\end{equation}
with ANN parameters $\theta$.
The vanilla \textit{Growing NCA} architecture---which we follow here---implements a 2D grid of $N=M\times M$ localized cellular agents that are equipped with the same feed-forward ANN:
Every cell $i$ filters its $3\times3$ Moore neighborhood $\mathcal{N}_i$ with a predefined convolutional perception kernel $\boldsymbol{p}_i^t = f_p(\boldsymbol{X}_i^t)$, followed by a dense hidden layer $\boldsymbol{h}_i^t=f_h(\boldsymbol{p}_i^t)$ with ReLU activation, and a linear output layer $\Delta\boldsymbol{x}_i^t=f_x(\boldsymbol{h}_i^t)$, such that $f_\theta=f_x \circ f_h \circ f_p$.
The perception kernel comprises an identity map $I(\cdot)$ and horizontal and vertical Sobel filters $S_x(\cdot)$ and $S_y(\cdot)$ of the local neighborhood $\mathcal{N}_i$, hence
$f_p(\cdot):\mathbb R^{(3\times 3)\times C}\mapsto\mathbb{R}^{3C}$,
$f_h(\cdot):\mathbb R^{3C}\mapsto\mathbb{R}^{H}$, and
$f_x(\cdot):\mathbb R^{H}\mapsto\mathbb{R}^{C}$.
The output layer $f_x(\cdot)$ is subjected to a stochastic binary update mask $m_i^t$, a layer that determines whether cell $i$ is updated at time $t$ during both training and inference.
This renders the individual cell-state updates asynchronous and introduces uncertainty and noise into the system.\cite{mordvintsev_growing_2020}

The cellular agents' ANN parameters $\theta$ can be optimized such that their shared policy, i.e., $f_\theta(\cdot)$, allows them to self-organize into a system-level configuration $\boldsymbol{X}^{T}=\{\boldsymbol{x}_i^{T}\}_{i\in\{1,\dotsc N\}}$ that resembles a predefined target pattern $\boldsymbol{Y}=\{\boldsymbol{y}_i\}_{i\in \{1,\dotsc N\}}$ on the $M\times M$  grid---such as an RGBA emoji---within a developmental time frame $T\sim\mathcal{U}\{T_\mathrm{D}\pm\Delta_\mathrm{D}\}$.
Target-specific parameters are obtained as 
\begin{equation}
\theta_{\boldsymbol{Y}}=\arg\min_{\theta}\;
    \mathbb{E}_{T\sim\mathcal{U},m_i^t}\left[
        ||\tilde{\boldsymbol{X}}^{T} - \boldsymbol{Y}||_F^2
    \right],
\label{eq:nca:loss}
\end{equation}
where the NCA evolves from an initial seed state $\boldsymbol{X}^{0}=\boldsymbol{X}_{\mathrm{seed}}$---typically from, but not restricted to a single cell at the center of the grid---before its deviation to the target pattern $\boldsymbol{Y}$ is evaluated at time $T$; $\tilde{\boldsymbol{X}}^{T}$ denotes the visible, i.e., RGBA channels of the complete grid state.

NCAs exhibit strong architectural parallels---i.e., inductive biases---with the agential substrate of biological matter.\cite{Hartl2025NCAs}
Although having important evolutionary  implications,\cite{Hartl2024MCA, Pande2023, bielawski_evolving_2024, pio2023scaling} optimization is most reliable with backpropagation training.\cite{mordvintsev_growing_2020}
Applications range from \textit{in silico} morphogenesis,\cite{mordvintsev_growing_2020} adversarial attacks~\cite{Randazzo2021, Cavuoti2022} and aging,~\cite{PioLopez2025} to self-organizing textures,\cite{Niklasson2021} and maze-solving~\cite{umu1729_maze_solver_2023} over self-classifying architectures~\cite{Randazzo2020} and sensorimotor tasks,\cite{najarro2022hypernca, Hartl2025MS, lopez2026BraiNCA} to abstract reasoning,\cite{guichard2025arcnca}.
The geometry and mechanistic interpretability of the regulatory parameter space of NCAs nevertheless remain largely unexplored; see Ref.~[\citenum{Hartl2025NCAs}] for a review.

\subsection{Low-Rank Adaptation of Neural Cellular Automata}
\label{sec:methods:lora}
% MULTI LAYER
NCAs can be composed of $l=1,\dotsc,L$ hidden layers of dimension $H_l$, such that $\boldsymbol{h}_{i, l}^t=f_h^{(l)}\left(\boldsymbol{h}_{i, l-1}^t\right)$, where the input to the first hidden layer $l=1$ is the perception vector, $\boldsymbol{h}_{i,0}^t=\boldsymbol{p}_i^t$, and the last hidden layer $L$ represents the output layer, $f_x(\cdot)\equiv f_h^{(L)}(\cdot)$.
% ACTIVATION
Each layer may be subject to non-linear activation or other filters, which we formally contract as $a_h^{(l)}(\dotsc)$ after signal aggregation, and define the layer $l$-specific activation as
\begin{equation}
    f_h^{(l)}\left(\boldsymbol{h}_{i, l-1}^t\right)=a_h^{(l)}\left(
        \boldsymbol{W}^{(l)}\;\boldsymbol{h}_{i,l-1}^t + \boldsymbol{b}^{(l)}
    \right),
\end{equation}
with all weights $\boldsymbol{W}^{(l)}\in\mathbb R^{H_l\times H_{l-1}}$ and biases $\boldsymbol{b}^{(l)}\in\mathbb{R}^{H_l}$ comprising the NCA's parameters $\theta$.

\begin{figure}
    \centering
    \includegraphics[width=\linewidth]{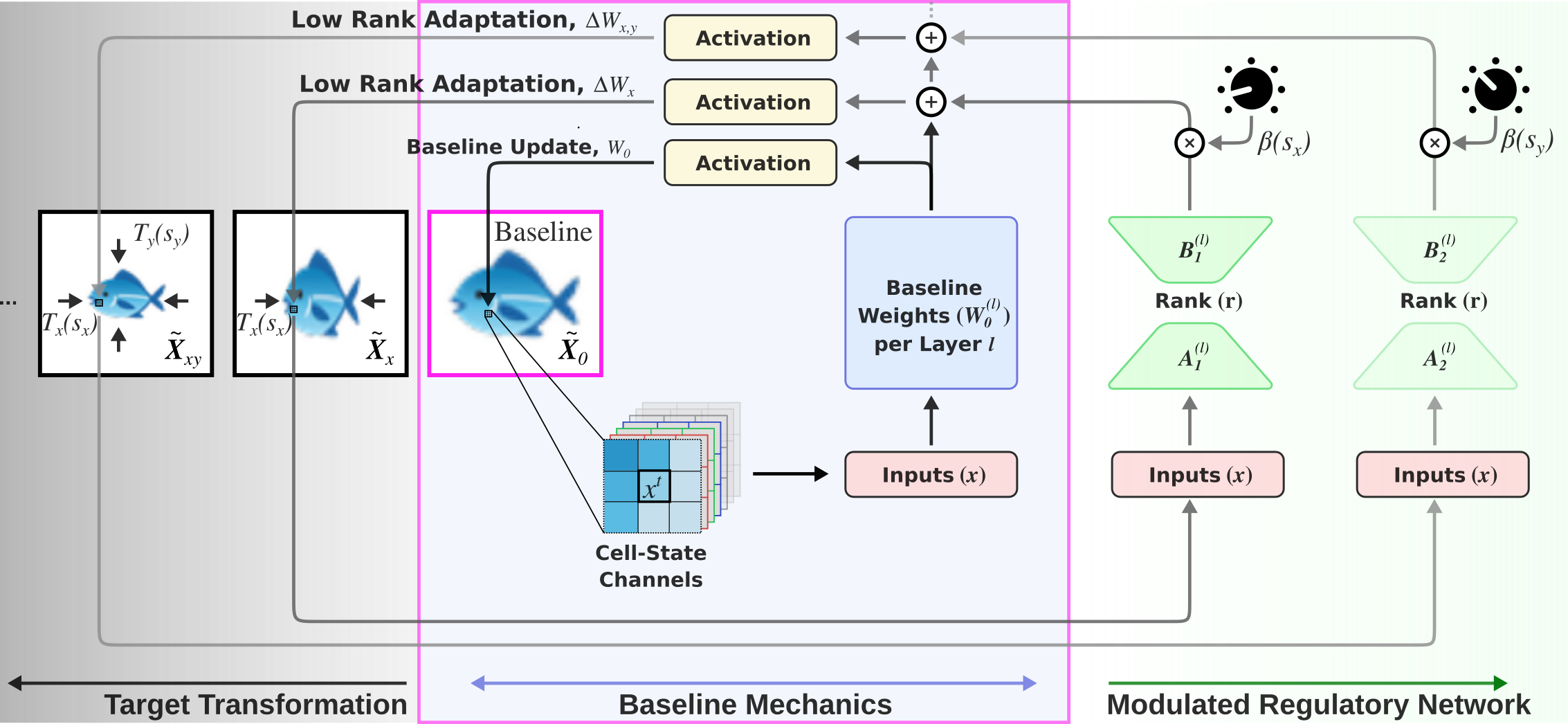}
    \caption{
        \textbf{Schematic illustration of mapping composite phenotypic transformations to mechanistic modulations in an NCA's regulatory network via LoRA.}
        We perform phenotypic variations $\boldsymbol{Y}\mapsto \boldsymbol{Y}_{xy}=\mathcal{T}_{xy}(\boldsymbol{s})\boldsymbol{Y}$ via parametric transformations $\mathcal{T}_{xy}\left(\boldsymbol{s}=\{s_x,s_y\}\right)$ and ask which LoRA parameter modulations $\theta_{xy}(\boldsymbol{s})= \theta + \Delta\theta_{xy}(\boldsymbol{s})$ give rise to modulated system-level dynamics $\tilde{\boldsymbol{X}}^{t}_{xy}(\boldsymbol{s})$ of a baseline NCA that resemble the transformed phenotype $\tilde{\boldsymbol{X}}^{T}_{xy}(\boldsymbol{s})\approx\mathcal{T}_{xy}(\boldsymbol{s})\boldsymbol{Y}$ after $T$ development steps.
    }
    \label{fig:lora:transform:model}
\end{figure}

Let $\theta_0^{(l)}=\{(\boldsymbol{W}_0^{(l)}, \boldsymbol{b}_0^{(l)})\}$ be the frozen parameter set of a reference frame network, e.g., of a pretrained NCA.
LoRA allows to refine---or fine-tune---the model's baseline behavior $\boldsymbol{Y}=\boldsymbol{Y}_0$ toward a new target $\boldsymbol{Y}_\mathcal{T}$ by modulating the frozen weights $\boldsymbol{W}_0^{(l)}$ via
\begin{equation}
    \boldsymbol{W}_{\mathcal{T}}^{(l)}=\boldsymbol{W}_0^{(l)} + \Delta\boldsymbol{W}_{\mathcal{T}}^{(l)},
    \quad
    \Delta\boldsymbol{W}_{\mathcal{T}}^{(l)}=\sum_{k=1}^{N_\mathcal{T}}\beta^{(l,r)}_k\boldsymbol{A}^{(l, r)}_k\boldsymbol{B}^{(l, r)}_k,
    \label{eq:lora:transform}
\end{equation}
i.e., by a composition of $k=1,\dotsc,N_\mathcal{T}$ rank $r$ factors $\boldsymbol{A}^{(l,r)}_k\in\mathbb R^{H_l,r}$ and $\boldsymbol{B}^{(l,r)}_k\in\mathbb R^{r, H_{l-1}}$ weighted by a trainable or predefined scaling factor $\beta_k^{(l,r)}\in\mathbb R$ that may or may not depend on $l$, $r$, or $k$.
Thus, when finetuning an NCA's behavior from $\boldsymbol{Y}_0\mapsto\boldsymbol{Y}_\mathcal{T}$, we only optimize $\boldsymbol{A}^{(l,r)}$ and $\boldsymbol{B}^{(l,r)}$ with $\mathcal{O}\left(N_\mathcal{T}(H_{l-1}+ H_l)r\right)$ parameters in \Cref{eq:nca:loss} instead of modulating all $\mathcal{O}\left(H_{l-1}\times H_l\right)$ parameters per layer.

Unless explicitly necessary, we drop layer and rank labels $(l,r)$ below for simplicity when referring to parameters across layers $\boldsymbol{W}_k\equiv\{\boldsymbol{W}_k^{(l)}\}_{l=1}^L$ and a specific rank, $\boldsymbol{A}_k, \boldsymbol{B}_k\equiv\{\boldsymbol{A}_k^{(l,r)},\boldsymbol{B}_k^{(l,r)}\}_{l=1}^{L}$.
Throughout the manuscript, the weights of both hidden and output layers are modulated with corresponding LoRAs of the same rank $r$.
Although we here assume a fixed perception block, e.g., trainable convolutional or attention-based kernels can also be modified by corresponding LoRAs.

%%%%%%%%%%%%%%%%%%%%%%%%%%%%%%%%%%%%%%%%%%%%%%%%%%%%%%%%%%%%%%%%%%%%%%%%%%%%%%%%%%%%%%%%%
%%% Results %%%%%%%%%%%%%%%%%%%%%%%%%%%%%%%%%%%%%%%%%%%%%%%%%%%%%%%%%%%%%%%%%%%%%%%%%%%%%
%%%%%%%%%%%%%%%%%%%%%%%%%%%%%%%%%%%%%%%%%%%%%%%%%%%%%%%%%%%%%%%%%%%%%%%%%%%%%%%%%%%%%%%%%
\section{Results}

\subsection{LoRA-NCA: Finetuning Pretrained Neural Cellular Automata}

We use Growing NCAs\cite{mordvintsev_growing_2020,Hartl2025NCAs} as minimal models of distributed bio-inspired development to study how changes of the interaction parameters at the intracellular self-regulatory level (akin to gene-regulatory modulations) translate into changes in collective morphology (e.g., Thompson's transformations). An NCA comprises an $M\times M$ grid of artificial cells (here $67\times 67$), each carrying a numerical state (here real-valued vectors of dimension $C=32$) and applying the same ANN as its update rule (the same $L=2$-layer ANN per cell with one hidden layer with $H_1=96$ neurons and one output layer with $H_2=C=32$ channels, amounting to $\approx12,416$ weights and bias parameters $\theta_0$; see \Cref{sec:methods:nca}). At each discrete developmental step, every cell perceives the state information from neighboring cells and updates its own state accordingly. This shared regulatory ANN is trained so that, through repeated local interactions, the system grows and maintains a target morphology from a single seed cell. The cellular states thereby provide an abstract representation of cellular physiology (such as transcriptomic expressions or cell-differentiation), while the shared ANN represents the regulatory machinery coordinating development. Although the regulatory machinery is learned rather than being specified \textit{a priori}, such NCAs' collective behavior is typically brittle to uncoordinated changes in their regulatory ANN parameters.\cite{Hartl2025NCAs}

After pretraining an NCA to grow a baseline morphology, we retain its ANN parameters $\theta_0$ as a ``reference'' regulatory program and study the effects of minimal subsequent adaptations. 
Specifically, we ask whether compact weight modulations can systematically impose functionally persistent and parametrically controllable transformations through development, and whether these adaptations transfer across morphologies: we specifically ask whether D'Arcy Thompson's shape transformations can be represented in a cybernetic tissue, i.e., by transferable low-dimensional directions in NCA weight space.

In this contribution, we identify Low-Rank Adaptation\cite{Edward2022LoRA} (LoRA)---a technique heavily used to fine-tune large language models---as an efficient approach to modulating an existing NCA's parameters such as to capture and parametrically control variations of the system-level outcomes.
In \Cref{sec:transform:xy} we demonstrate that a single rank $r=1$ modulation---comprising $320$ parameters, i.e., only $2.6\%$ of the NCA's full ANN parameter count---is capable of parametrically controlling the effective size of fully grown phenotypes through modified developmental pathways.
In \Cref{sec:transform:phenotype} we test whether these trained scale-transforming rank-one LoRAs transfer across different NCAs (other phenotypes) and empirically identify in \Cref{sec:lora:pca} degenerate regulatory programs---weight modulations---that for instance control scale transformations, stylistic features, and induce phenotypic ``fission'' from a large dataset of $\approx 25{,}000$ LoRA-NCAs of rank $r=16$ that share the same scaffold $\boldsymbol{W}_0$; for further simulation and training details refer to \Cref{app:param}.

%%%%%%%%%%%%%%%%%%%%%%%%%%%%%%%%%%%%%%%%%%%%%%%%%%%%%%%%%%%%%%%%%%%%%%%%%%%%%%%%%%%%%%%%%
%%% Results: Mapping Trafo. with Weight-Modulation %%%%%%%%%%%%%%%%%%%%%%%%%%%%%%%%%%%%%%
%%%%%%%%%%%%%%%%%%%%%%%%%%%%%%%%%%%%%%%%%%%%%%%%%%%%%%%%%%%%%%%%%%%%%%%%%%%%%%%%%%%%%%%%%
\subsection{Learned Regulatory Hyper-Directions control Phenotypic Scale-Variation}
\label{sec:transform:xy}
% GOAL
It is a long-standing question whether phenotypic variations under specific spatial transformations can be implemented mechanistically, i.e., how transformation-conditional modulations of an organism's regulatory networks can induce altered developmental outcomes that implement the transformed phenotype.
% Addressing this in biophysically realistic---yet alone biological---models is still out of scope.
In biological terms, this means identifying coordinated changes in cellular gene-regulatory and biophysical networks that alter how cells grow, differentiate, move, and communicate during development, such that their collective behavior generates the transformed anatomy.
To address this, we here rely on the \textit{in silico} growth dynamics implemented by NCAs. 
% IMPLEMENTATION
Specifically, we ask how phenotypic variations $\boldsymbol{Y}_\mathcal{T}\mapsto\mathcal{T}(\boldsymbol{s}) \boldsymbol{Y}$ under parametric transformations $\mathcal{T}(\boldsymbol{s})$ map to adaptations of the ANN parameters $\theta_\mathcal{T} \mapsto  \theta + \Delta\theta_{\mathcal{T}}(\boldsymbol{s})$ that give rise to modulated system-level dynamics ${\boldsymbol{X}}^t_{\mathcal{T}}$ that resemble the transformed phenotype $\tilde{\boldsymbol{X}}^T_{\mathcal{T}}(\boldsymbol{s})\approx\mathcal{T}(\boldsymbol{s})\boldsymbol{Y}$ after development $T\sim\mathcal{U}\{T_D\pm\Delta_D\}$.
Here, $\boldsymbol{s}$ parametrizes a morphological transformation (akin, but not limited, to Thompson's grid transformations), while the NCA parameter modulations represent corresponding interventions at the GRN level that would causally explain the macroscopic transformations through modulated developmental pathways.

Let $\mathcal{T}(\boldsymbol{s})=\{T_k(s_k)\}_{k=1}^{N_\mathcal{T}}$ be a set of spatial transformations $T_k(s_k)$, parametrized by $\boldsymbol{s}=\{s_k\}_{k=1}^{N_\mathcal{T}}$, that iteratively transform the shape of a baseline phenotype $\boldsymbol{Y}=\boldsymbol{Y}_0$ via
\begin{equation}
    \boldsymbol{Y}_\mathcal{T}:\quad\boldsymbol{Y}_k(\boldsymbol{s})=T_{k}(s_k)\boldsymbol{Y}_{k-1}(\boldsymbol{s}).
    \label{eq:phenotype:transform}
\end{equation}
We map the compositional phenotypic transformation $\mathcal{T}(\boldsymbol{s})$ given by \Cref{eq:phenotype:transform} to parametrized modulations of the NCA's baseline weights ${\boldsymbol{W}}_{\mathcal{T}}^{(l)}(\boldsymbol{s})={\boldsymbol{W}}_0+\Delta{\boldsymbol{W}}_{\mathcal{T}}(\boldsymbol{s})$ by following \Cref{eq:lora:transform}, i.e., via LoRA decomposition $\Delta{\boldsymbol{W}}_{\mathcal{T}}(\boldsymbol{s})=\sum_{k=1}^{N_\mathcal{T}}\beta(s_k)\boldsymbol{A}_k\boldsymbol{B}_k$, see \Cref{fig:lora:transform:model}.
The corresponding rank $r$ adapters $\boldsymbol{A}_k$ and $\boldsymbol{B}_k$ can be trained to implement specific transformation $T_k(\cdot)$, and the linear coefficients $\beta(s_k)$ encode the strength at which the $T_k$-specific adaptations $\boldsymbol\Delta_k=\boldsymbol{A}_k\boldsymbol{B}_k$ modulate the scaffold $\boldsymbol{W}_{0}$.

\begin{figure}
    \centering
    \includegraphics[width=\linewidth]{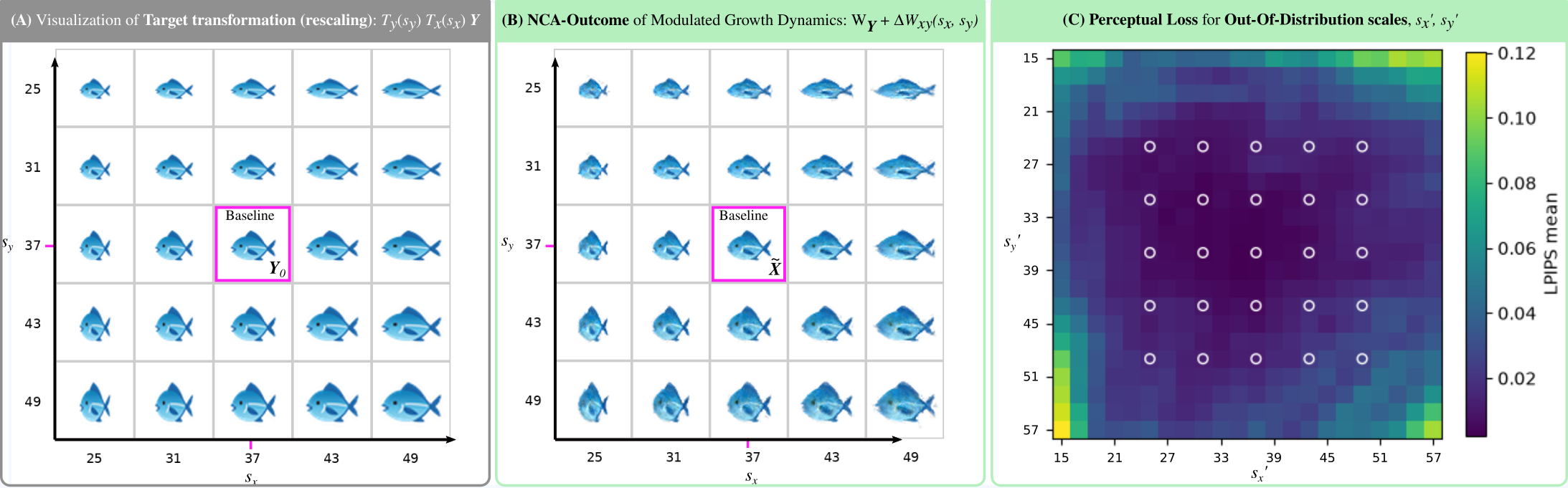}
    \caption{
        \textbf{Contrasting (A) target scaling transformation to (B) phenotypic outcomes of modulated NCA dynamics and (C) generalization experiments comparing out-of-distribution (ood) NCA modulations to expected target transformations.}
        (A): $xy$-scaling transformations $\boldsymbol{Y}_{xy}=T_y(s_y)T_x(s_x)\boldsymbol{Y}$ of a base fish emoji $\boldsymbol{Y}$ (magenta box) from $(s_0\times s_0)=(37x37)$ pixels to $25$ combinations of $(sx\times s_y)$ pixels with $s_x,s_y\in\{25, 31, 37, 43, 49\}$.
        (B): Exemplary outcomes $\tilde{\boldsymbol{X}}_{xy}^{T_D}$ of the developmental process of $(s_x,s_y)$-parametrically modulated NCAs, see \Cref{eq:lora:transform:xy}, after $t=T_D$ steps.
        (C): Heatmap of the mean perceptual loss~\cite{Zhang2018PerceptualLoss} (LPIPS, see \Cref{app:metrics:perceptual:loss}) comparing the phenotypic outcomes of ood-modulated NCAs to the expected but withheld targets across a fine grid of scale transformations $s_x^\prime,s_y^\prime\in[15,57]$; we used $16$ independent stochastic NCA rollouts per scaling and refer to \Cref{app:fig:lora:generalize:xy} for more details.
    }
    \label{fig:lora:transform:xy}
\end{figure}

% EXAMPLE TARGET XY SCALING
Here, we apply orthogonal scaling transformations $T_1\equiv T_x$ and $T_2\equiv T_y$ into horizontal and vertical directions, $x$ and $y$, to a baseline emoji phenotype $\boldsymbol{Y}_0$, see \Cref{fig:lora:transform:xy}~A. This $xy$-scaling transformation is parametrized by $s_1=d_x\,s_0$ and $s_2=d_y\,s_0$, such that $d_k\neq 1$ scales the phenotype by a factor of $d_k$ along direction $k$, while the baseline is recovered for $s_k=s_0$.

% EXAMPLE MECHANISTIC MAPPING
For the mechanistic LoRA mapping, we chose $\beta_k^{(l,r)}=\beta(s_k)=\beta_0\log(s_k/s_0)$ in \Cref{eq:lora:transform} such that reciprocal scaling is symmetric, $d_k^{\pm 1}s_0\mapsto\pm\beta_0\log d_k$, and $s_k=s_0\mapsto\beta_k=0$ leaves the NCA's parameters unchanged; $\beta_0$ is a constant.
The weight modulation for $xy$-scaling becomes
\begin{equation}
    \Delta\boldsymbol{W}_{xy}(d_x, d_y)=\beta_0\sum_{k\in\{x,y\}}\log(d_k)\,\boldsymbol{A}_k \boldsymbol{B}_k.
    \label{eq:lora:transform:xy}
\end{equation}

% EXAMPLE TRAINING
Assuming an NCA with frozen parameters $\theta_{0}$ that implements a developmental process of a baseline phenotype $\tilde{\boldsymbol{X}}^{T}\approx\boldsymbol{Y}_0$, we optimize the LoRAs $\boldsymbol{A}_k$, $\boldsymbol{B}_k$ in \Cref{eq:lora:transform:xy} such that the  modulated outcome matches the transformed target by minimizing $\mathbb E_{T,\boldsymbol{s}}\left[||\tilde{\boldsymbol{X}}^T_{xy}(\boldsymbol{s})-\boldsymbol{Y}_{xy}(\boldsymbol{s})||^2\right]$.

% RESULTS
This approach turns out highly effective: 
In \Cref{fig:lora:transform:xy}~A, we scaled a baseline fish emoji of size $(s_0\times s_0)=(37\times37)$ pixels to permutations of $(s_x\times s_y)$ pixels $s_x, s_y\in\{25, 31, 37, 43, 49\}$.
% WORKS WITH r=1 !!
The mapped NCA outcomes depicted in \Cref{fig:lora:transform:xy}~B demonstrate that even rank $r=1$ adaptations of the NCA's baseline weights, i.e., $\boldsymbol A_k^{(l)}\in\mathbb{R}^{H_l\times 1}$ and $\boldsymbol B_k^{(l)}\in\mathbb{R}^{1\times H_{l-1}}$, suffice to robustly implement the modulated developmental pathways for all $25$ investigated scaling transformations.
%GENERALIZES TO UNSEEN sx AND sy VALUES
Moreover, the learned weight modulations, $\beta(s_x)\boldsymbol\Delta_{x}$ and $\beta(s_y)\boldsymbol\Delta_y$ generalize well to unseen phenotypic sizes $s_{x,y}^\prime\notin\{25, 31, 37, 43, 49\}$, cf. \Cref{fig:lora:generalize:xy}, and represent hyper-directions $\boldsymbol\Delta_{x,y}$ for modularly scaling the NCA's outcome along orthogonal $x$- and $y$-directions in phenotypic space by linear factors $\beta(s_{x,y})$.

LoRA is highly effective to systematically fine-tune---i.e., refine---the behavior of an existing NCA toward continuous phenotypic variations via low-rank weight modulations.
Moreover, we confirm that phenotypic transformations can be encoded in---and conditionally controlled by---low-rank hyper-directions of the regulatory networks of multiscale cybernetic substrates.
This suggests important implications for related biological systems (GRNs, bioelectric networks, etc.), biomedical applications (stabilizing system-level affectors like cancer, or aging), and, more generally, complex-systems-engineering~\cite{Levin2024MultiscaleWisdom}.
Yet, the nature, interpretability, and the potential for individualized vs. universal applicability of such hyper-directions remains elusive and requires extensions of the NCA framework to more realistic bio-physical models in future work.

%%%%%%%%%%%%%%%%%%%%%%%%%%%%%%%%%%%%%%%%%%%%%%%%%%%%%%%%%%%%%%%%%%%%%%%%%%%%%%%%%%%%%%%%%
%%% Results: Generalization of Scaling Hyper-Directions %%%%%%%%%%%%%%%%%%%%%%%%%%%%%%%%%
%%%%%%%%%%%%%%%%%%%%%%%%%%%%%%%%%%%%%%%%%%%%%%%%%%%%%%%%%%%%%%%%%%%%%%%%%%%%%%%%%%%%%%%%%
\subsection{LoRA finds Universal Mechanistic Hyper-Directions of Phenotypic Variation, in Emoji-World}
\label{sec:transform:phenotype}
The above ``hyper-directions of scale'', i.e., $\boldsymbol\Delta_{x,y}$, generalize well to weight-space modulations across unseen $xy$-transformations $\boldsymbol{s}_{x,y}^\prime$ of the same baseline NCA, see \Cref{fig:lora:transform:xy}~C.
Yet, we can't expect \textit{ad hoc} transfer to other NCAs that have been independently trained on different targets $\boldsymbol{X}_k^T\approx\boldsymbol{Y}_k$. 
Due to initial conditions, or pure chance, single weights or entire channels might be randomly permuted or subject to other gauge symmetries. 
Thus, two independently trained NCAs that implement the exact same dynamics most likely obtain mechanistically incompatible weights.

\begin{figure}
    \centering
    \includegraphics[width=0.75\linewidth]{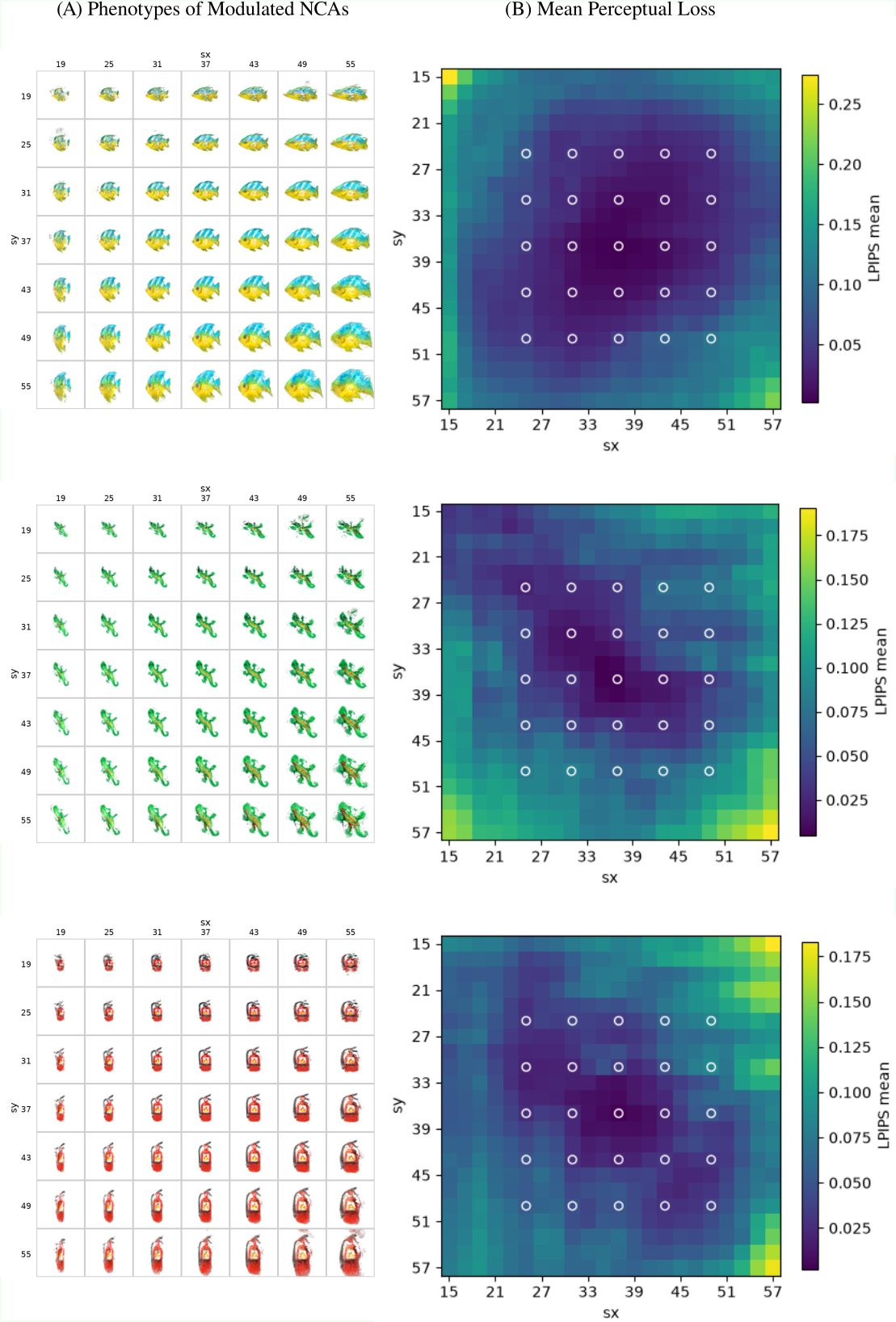}
    \caption{
        \textbf{Low-rank regulatory adaptations learned to rescale one emoji transfer zero-shot to diverse phenotypes.}
        LoRAs trained on a discrete set of $xy$-scaling transformations for a single phenotype $\boldsymbol{Y}$---with $s_x,s_y\in\{25, 31, 37, 43, 49\}$, see \Cref{fig:lora:transform:xy}---not only generalize zero-shot to out-of-distribution scaling $s_x^\prime,s_y^\prime$, but also transfer across phenotypes, i.e., to NCAs with shared reference scaffold $\boldsymbol{W}_0$ but fine-tuned on a different target $\boldsymbol{Y}_\phi$.
        Column (A) shows developmental outcomes for three different targets across selected sizes $s_x^\prime,s_y^\prime\in\{17, 25, 31, 37, 43, 49, 55\}$---from top to bottom:
        a similar fish phenotype to the original blue-fish emoji in \Cref{fig:lora:transform:xy}, but with different color-composition and internal features;
        completely different green lizard phenotype;
        and a red fire extinguisher.
        Column (B) depicts the mean perceptual loss~\cite{Zhang2018PerceptualLoss} (LPIPS, see \Cref{app:metrics:perceptual:loss}) per emoji (row) across a fine grid of scale transformations $s_x^\prime,s_y^\prime\in[15,57]$; we used $16$ independent stochastic NCA rollouts per ($\phi$, $s_x^\prime$,$s_y^\prime$) tuple; refer to \Cref{app:fig:lora:generalize:xy} for the corresponding perceptual loss STD and MSE loss; the latter is, however, mostly correlated with the overall phenotypic size rather than actual pixel-wise or semantic accuracy of the modulated developmental outcomes.         
        Especially from the perceptual loss (B) we learn that all of the tested target phenotypes $\phi$ can be scaled via the same hyper-directions $\boldsymbol\Delta_{x,y}$, even beyond training conditions; the white circles in (B) emphasize training conditions, where training has only been conducted on the blue fish target shown in \Cref{fig:lora:transform:xy}.
        This strongly suggests that these LoRAs are universal directions of scale for NCAs with shared scaffold.
        }
    \label{fig:lora:generalize:xy}
\end{figure}

However, we can fine-tune a shared baseline NCA with parameters $\theta_0=\{(\boldsymbol{W}_0, \boldsymbol{b}_0)\}$ on different targets $\boldsymbol{Y}_k$ by jointly training a set of independent, i.e., target $\boldsymbol{Y}_k$-specific LoRAs, $\{\boldsymbol{A}_k, \boldsymbol{B}_k\}_k$.
To grow a specific target $\tilde{\boldsymbol{X}}^T(k^\prime)\approx\boldsymbol{Y}_{k^\prime}$, we activate a single corresponding weight modulation, $\Delta W_{k^\prime}=\boldsymbol{A}_{k^\prime}\boldsymbol{B}_{k^\prime}$, using a Kronecker-Delta (one-hot) encoding for $\beta_k=\delta_{kk^\prime}$ in \Cref{eq:lora:transform}.
We can simultaneously learn a shared baseline $\boldsymbol{W}_0$ and target-specific adapters $\{\boldsymbol{A}_k \boldsymbol{B}_k\}_k$ across a wide range of $N_\phi$ targets by minimizing $\mathbb{E}_{T}=\frac{1}{N_\phi}\sum_{k^\prime=1}^{N_\phi} ||\tilde{\boldsymbol{X}}^{T}(k^\prime) - \boldsymbol{Y}_{k^\prime}||^2$ under joint variation of the scaffold parameters $\theta_0$ and all target-specific adapters $\{(\boldsymbol{A}_k^{(l,r)}, B_k^{(l,r)})\}_{k, l}$ across all layers $l$.

In fact, the NCA shown in \Cref{fig:lora:transform:xy}, let its target-label be $k=1$, has been trained jointly with $N_\phi=120$ different, randomly chosen target emojis.
Thus, its target-specific weights can be decomposed into $\boldsymbol{W}_{1}=\boldsymbol{W}_{0}+\Delta\boldsymbol{W}_{k^\prime=1}$, i.e., a shared scaffold $\boldsymbol{W}_0$ and a target-specific modulation $\Delta\boldsymbol{W}_{k^\prime=1}=\boldsymbol{A}_{k^\prime}\boldsymbol{B}_{k^\prime}$.
We report that all other $N_\phi$ targets have been trained successfully, yielding a set of $120$ NCA LoRAs $\{\boldsymbol{A}_{k}\boldsymbol{B}_k\}_{k=1}^{N_\phi}$ of rank $r=16$ with a single shared weight-space reference frame $\theta_0=\{\boldsymbol{W}_0,\boldsymbol{b}_0\}$ that has no explicit phenotypic target.

In turn, the experiments discussed in \Cref{sec:transform:xy} suggest that LoRAs can be stacked efficiently, i.e., not only covering continuous transformations $T_x$ and $T_y$, as in \Cref{eq:lora:transform:xy}, but also covering tailored phenotypic modulations: 
the corresponding NCA's regulatory network is simultaneously composed of discrete phenotypic modulations of rank $16$, $\Delta W_{k^\prime=1}$, and continuous parametric modulations of rank $1$, i.e., $\beta(s_q)\boldsymbol\Delta_q$, with the transformation-specific weights $\boldsymbol W_{xy}(k^\prime=1;s_x, s_y)=\boldsymbol{W}_0+\Delta\boldsymbol{W}_{k^\prime}+\sum_{q\in\{x,y\}}\beta(s_q)\boldsymbol\Delta_q$.

Moreover---and strikingly---although the $xy$-scaling transformations $\beta(s_{x,y})\boldsymbol\Delta_{x,y}$ have been obtained for a single phenotype $k^\prime=1$, they apply zero-shot to all tested phenotypes with shared scaffold $\boldsymbol{W}_0$---without finetuning to any other target $k\neq k^\prime$.
\Cref{fig:lora:generalize:xy} demonstrates this qualitatively on three targets that are vastly different in structure, orientation, color, and semantic features compared to the baseline blue fish emoji depicted in \Cref{fig:lora:transform:xy}:
While global shapes transform parametrically according to $s_x$ and $s_y$, and even generalize to out-of-distribution conditions $s_x^\prime,s_y^\prime\notin\{25, 31, 37, 43, 49\}$, phenotype-specific internal features and proportions are preserved exceptionally well across all tested targets, with anisotropic scaling additionally producing effective shear transformations for oriented phenotypes.

That strongly suggests the $\boldsymbol\Delta_{x,y}$ adapters learned something close to system-level geometric control, rather than memorizing particular features of the baseline phenotype.
That is, the adapters are actually behaving compositionally.
We conclude that the $xy$-scaling modulations $\boldsymbol\Delta_x$ and $\boldsymbol\Delta_y$ represent ``universal hyper-directions of scale'' across the dataset of target emojis, and we can write a closed-form $xy$-scaling transformation for all $k=1,\dotsc,N_\phi$ phenotypes
\begin{equation}
    \boldsymbol{W}_{xy}(k^\prime;s_x, s_y)=
        \boldsymbol{W}_0+
        \sum_{k=1}^{N_\phi}\delta_{k^\prime k}\Delta\boldsymbol{W}_{k}+
        \sum_{q\in\{x,y\}}\beta(s_q)\boldsymbol\Delta_q.
\end{equation}

Within the present framework, D'Arcy Thompson's grid transformations can thus be understood as low-dimensional latent variables in the NCA's parameter space, i.e., hyper-directions in the self-regulatory networks that implement---and parametrically control---these transformations mechanistically.
The shared scaffold $\boldsymbol{W}_0$ not only represents a shared reference frame in weight space for the different NCAs but allows efficient transfer learning via mechanistic hyper-directions across a broad range of discrete and continuous phenotypic variations. 

Although many questions remain, this raises exciting potential for future biomedical applications: if, for instance, universal hyper-directions responsible for the regenerative capabilities of certain species can be found in their GRNs or bioelectric networks (e.g., with deep-learning approaches of transcriptomic data), 
this could potentially provide intervention targets for developmental and regenerative control.
However, first efforts to find a weight-space mapping between the regulatory machinery of non-regenerative to regenerative NCAs did not yield a generalizable hyper-direction---for details refer to \Cref{app:lora:regeneration}.
Future work will be dedicated to exploring this option further, especially how to transfer regenerative capabilities from regenerative NCA solutions to those without regenerative abilities.

%%%%%%%%%%%%%%%%%%%%%%%%%%%%%%%%%%%%%%%%%%%%%%%%%%%%%%%%%%%%%%%%%%%%%%%%%%%%%%%%%%%%%%%%
%%% Results: PCA %%%%%%%%%%%%%%%%%%%%%%%%%%%%%%%%%%%%%%%%%%%%%%%%%%%%%%%%%%%%%%%%%%%%%%%%
%%%%%%%%%%%%%%%%%%%%%%%%%%%%%%%%%%%%%%%%%%%%%%%%%%%%%%%%%%%%%%%%%%%%%%%%%%%%%%%%%%%%%%%%%
\subsection{Phenotypic Variation is encoded in Latent Regulatory Geometry}
\label{sec:lora:pca}
Above, we learn an imposed mapping between parametric transformations of a baseline phenotype and corresponding weight adaptations to a reference NCA that modulate the developmental dynamics toward the desired transformed outcomes.
Given that LoRA is efficient to find weight adaptations for much more general phenotypic variations---qualitatively different shapes, colors, and styles---we here ask whether a comprehensive empirical dataset of LoRAs captures latent semantic hyper-directions relevant to developmental outcomes.

\begin{figure}
    \centering
    \includegraphics[width=\linewidth]{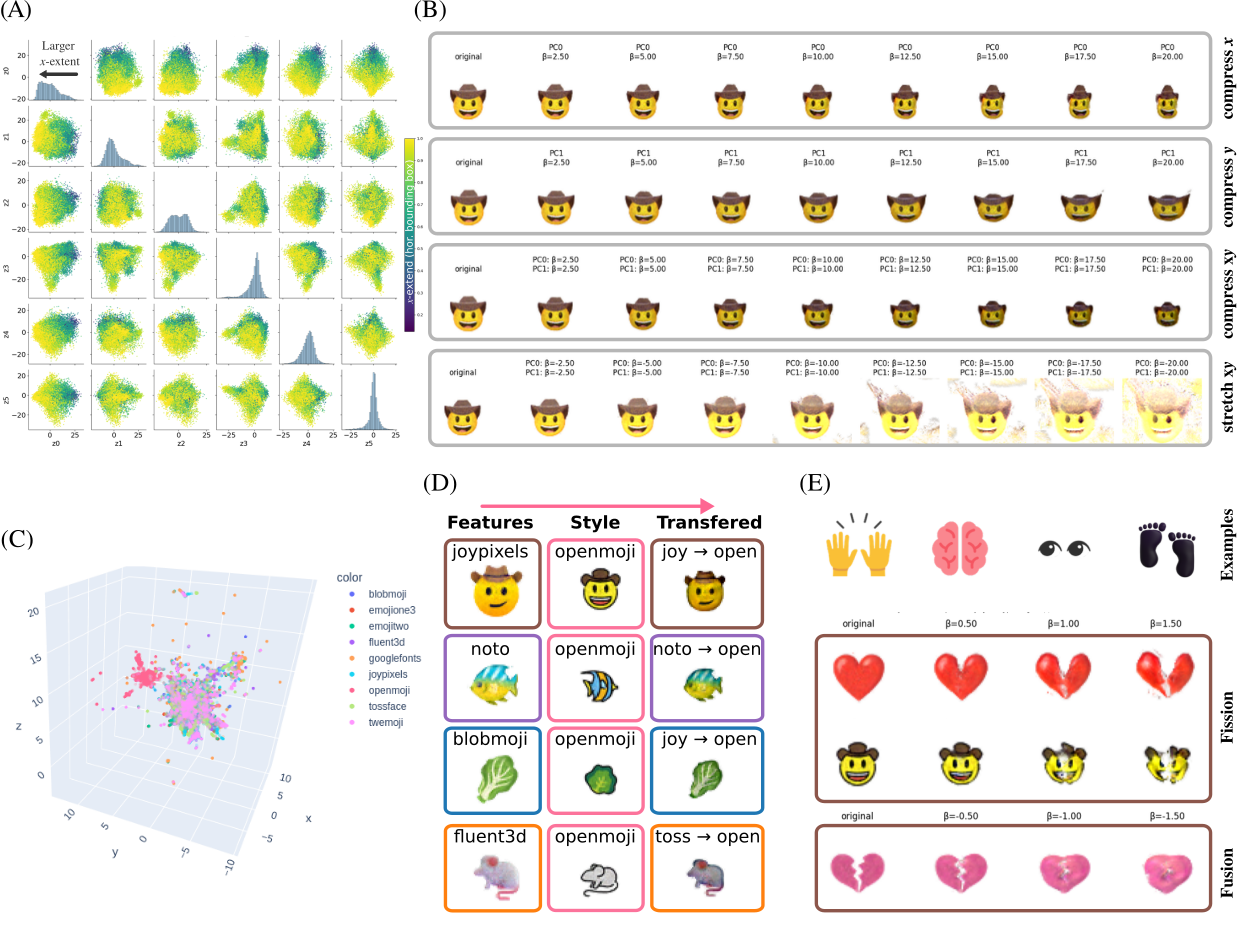}
    \caption{
        \textbf{Weight-Space Principal Component Analysis (PCA) reveals Semantic Hyper-Directions across NCA Growth-Dynamics.}
        (A) Projection of full adapter weight modulations, $\Delta \boldsymbol{W}_\phi=\boldsymbol{A}_\phi \boldsymbol{B}_\phi$, onto leading six principal components, $PC_{0\dotsc 5}$, colored by the relative horizontal ($x$) extent of the corresponding target emojis $\boldsymbol{Y}_\phi$. 
        The first two principal components, $PC_0$ and $PC_1$, strongly correlate with horizontal ($x$) and vertical ($y$) phenotypic extent, respectively ($r(PC_0,x)=-0.71$ and $r(PC_1,y)=-0.73$, measured from the targets' bounding boxes).
        (B) In the top (second-to-top) panel, the growth-dynamics of a baseline cowboy-hat NCA is compressed into $x$ ($y$) direction by adding $\beta\,\Delta\boldsymbol{W}_{PC_0}$ ($\beta\,\Delta\boldsymbol{W}_{PC_1}$) to its phenotype-specific weight modulation $\Delta\boldsymbol{W}_\phi$, where $\Delta\boldsymbol{W}_{PC_i}$ denotes the corresponding principal direction mapped from standardized PCA coordinates back to the original weight coordinates. 
        In the second-to-bottom (bottom) panel, we shrink (stretch) the growth-dynamics of the same NCA into $x$ and $y$ directions simultaneously by adding (subtracting) $\beta\,(\Delta\boldsymbol{W}_{PC_0} + \Delta\boldsymbol{W}_{PC_1})$ to (from) baseline weights. 
        Unlike the universal scaling directions $\boldsymbol\Delta_{x,y}$, the principal components $PC_{0,1}$ also affect the brightness of the modulated NCAs' outcomes.
        (C) UMAP projection~\cite{McInnes2018umap} of the leading 50 PCs from (A); coloring by vendor, and especially the separation of the ``OpenMoji'' style (salmon) from the remaining vendors reveals that the weight-space captures style-specific information. 
        (D) The mean weight-space translation vector $\Delta\boldsymbol{W}_\mathrm{open}=\langle\Delta\boldsymbol{W}_\phi\rangle_{\mathrm{open}}-\langle\Delta\boldsymbol{W}_\phi\rangle_{\mathrm{non-open}}$ from NCAs of any vendor (all but salmon in C) to OpenMoji NCAs (salmon in C) yields a style-transfer hyper-direction for OpenMoji-likeness.
        Four selected NCA experiments (from top to bottom) demonstrate that baseline features (left panels) are preserved by the OpenMoji style-transfer, but an OpenMoji-typical blackish outline (cf. center panels) is ``added'' to the NCAs' developmental dynamics (right panels).
        (E) A ``fission''-hyper-direction $\Delta\boldsymbol{W}_\mathrm{F}$ has been identified by subtracting the centroid of the remaining NCAs from that of the symmetrically split targets (see top-row examples), $\Delta\boldsymbol{W}_\mathrm{F}=\langle\Delta\boldsymbol{W}_\phi\rangle_{\mathrm{split}}-\langle\Delta\boldsymbol{W}_\phi\rangle_{\mathrm{non-split}}$, which can be used to parametrically split (fuse) baseline growth-dynamics along the vertical axis by adding (subtracting) $\beta \Delta\boldsymbol{W}_\mathrm{F}$ to (from) any baseline weights $W_\phi$. 
    }
    \label{fig:lora:pca}
\end{figure}

We thus train a vast dataset of LoRA NCAs, $\{\boldsymbol{A}_\phi,\boldsymbol{B}_\phi\}$, with the shared frozen scaffold $\boldsymbol{W}_0$ from \Cref{sec:transform:phenotype}, so the corresponding NCA's system-level outcome resembles the desired phenotype $\tilde{\boldsymbol{X}}^T_\phi\approx \boldsymbol{Y}_\phi$ out of $N_\phi\approx25,000$ targets (spanning vastly different shapes, colors, and stylistic features).
As shown in \Cref{fig:lora:pca}, principal component analysis~\cite{Hastie2009} (PCA) in weight space $\Delta\boldsymbol{W}_\phi=\boldsymbol{A}_\phi\boldsymbol{B}_\phi$---concatenating the layer matrices into a one-dimensional vector before standardizing the data to zero-mean and unit-variance---reveals semantically rich features in the empirically gathered dataset of regulatory NCA weights: some PCs strongly correlate with phenotypic features.
Notably, PCA in raw LoRA spaces yields limited information (see \Cref{app:lora:pca}).
Although each element of the dataset is obtained through a rank-16 LoRA, a PCA-derived axis captures a high-variance direction across many such adaptations in the full weight space and therefore need not correspond to a low-rank layer-wise update; movement along that axis nevertheless can provide a one-parameter control handle over specific phenotypic features.

The first and second principal components, $PC_0$ and $PC_1$, strongly correlate with horizontal, $x$, and vertical, $y$, phenotypic extent of the NCAs' targets (see \Cref{fig:lora:pca}~A,B), with Pearson coefficient $r(PC_0,x)=-0.71$ and $r(PC_1,y)=-0.73$, and cross-correlations  $r(PC_0,y)=0.07$ and $r(PC_1,x)=0.08$.
Modulating any phenotypic NCA's weight modulations $\Delta\boldsymbol{W}_{\phi}$ along the weight-space direction $\pm\beta\Delta\boldsymbol{W}_{PC_i}$ (corresponding to the first or second principal directions $i=0,1$) modularly compresses or stretches the NCA's growth-dynamics either in the $x$- or $y$-direction, simply via scaling the coefficient $\beta$.
Symmetrical modulations $W_\phi\pm\beta (\Delta\boldsymbol{W}_{PC_0}+\Delta\boldsymbol{W}_{PC_1})$ change the total size of the developmental outcome.
Interestingly, the empirically found scale directions, $\Delta\boldsymbol W_{PC_0}$ and $\Delta\boldsymbol W_{PC_1}$, only exhibit $\approx0.1$ cosine similarity with the corresponding learned scale hyper-directions, $\boldsymbol\Delta_x$ and $\boldsymbol\Delta_y$. 
This suggests a degeneracy of morphogenetic control, whereby distinct directions in parameter space can modulate the same macroscopic transformations---in a poly-computing sense: an agential substrate supports overlapping functional organizations along multiple paths~\cite{Edelman2001Degeneracy, bongard_2023_plentyofroom}.

Similarly, we report that the weight-space captures stylistic information (\Cref{fig:lora:pca}~C, D): 
Emojis of the OpenMoji vendor display a particular cartoonish style with a black outline, which is clearly separate from all other NCAs.
The mean translation vector in weight-space $\Delta\boldsymbol{W}_\mathrm{open}=\langle \Delta\boldsymbol{W}_\phi\rangle_{\mathrm{open}}-\langle \Delta\boldsymbol{W}_\phi\rangle_{\mathrm{non-open}}$ from NCAs that grow emojis of any other style yields a style-transfer hyper-direction toward ``OpenMoji-ness''.
Translating along this direction in weight-space $\Delta\boldsymbol{W}_\phi\pm\beta\Delta\boldsymbol{W}_\mathrm{open}$ thus aligns the style of a baseline NCA toward OpenMoji-ness---the typical blackish outline---while preserving other baseline features and details.

As a last example, we identified a non-trivial hyper-direction of vertical ``fission'', $\Delta\boldsymbol{W}_\mathrm{F}$:
The dataset comprises phenotypes consisting of multiple spatially separate clusters, e.g., of twins or mirror symmetric shapes; see examples in \Cref{fig:lora:pca}~E and \Cref{app:lora:pca:twinscore} for metric details.
Subtracting the centroid of the remaining NCAs from that grow such symmetrically split targets, $\Delta\boldsymbol{W}_\mathrm{F}=\langle\Delta\boldsymbol{W}_\phi\rangle_{\mathrm{split}}-\langle\Delta\boldsymbol{W}_\phi\rangle_{\mathrm{non-split}}$, yields a functional hyper-direction that, indeed, causes the developmental outcome of a baseline NCA  to split symmetrically along the vertical axis upon weight modulation $\Delta\boldsymbol{W}_\phi + \beta\Delta\boldsymbol{W}_\mathrm{F}$. 
Above a target-specific threshold, $\beta\gtrsim2$, the modulated NCA dynamics robustly grow and maintain split versions of the respective targets, preserving features in the respective spatial clusters.
Conversely, the inverse direction $\Delta\boldsymbol{W}_\phi - \beta\Delta\boldsymbol{W}_\mathrm{F}$ allows to fuse vertically split targets, although some details can be lost.

Given the correct basis---weight-space rather than LoRAs, in our case, see \Cref{app:lora:pca}---empirical regulatory data can yield surprisingly clear hyper-directions that linearly connect phenotypic features with regulatory encodings.
In other words, PCA does not merely organize adapters according to phenotype; movement along the inferred axes causally modulates the developmental outcome.
However, unlike the learned hyper-directions of scale $\boldsymbol\Delta_{x,y}$ discussed in \Cref{sec:transform:phenotype}, both $PC_0$ and $PC_1$, for instance, also affect the brightness of the corresponding developmental outcome (\Cref{fig:lora:pca}~B).
Analogously, style-transfer along $|\beta|\Delta\boldsymbol{W}_\mathrm{open}$ shrinks the modulated growth-dynamics to match the OpenMoji mean emoji size; i.e., the style-transfer simultaneously ``adds'' blackish outlines and shrinks the NCAs' outcomes.
Interestingly, $\Delta\boldsymbol{W}_\mathrm{open}$ is only weakly aligned with the scale-associated $PC_0$ and $PC_1$ directions (cosine similarity $\approx0.1$), despite producing a pronounced change in phenotype size.
Systematically extracting isolated, feature-specific hyper-directions from empirical data proves non-trivial, even for a structured space like ours, and motivates approaches capable of resolving nonlinear and/or entangled structure, as discussed next.

%%%%%%%%%%%%%%%%%%%%%%%%%%%%%%%%%%%%%%%%%%%%%%%%%%%%%%%%%%%%%%%%%%%%%%%%%%%
%%% Discussion %%%%%%%%%%%%%%%%%%%%%%%%%%%%%%%%%%%%%%%%%%%%%%%%%%%%%%%%%%%%
%%%%%%%%%%%%%%%%%%%%%%%%%%%%%%%%%%%%%%%%%%%%%%%%%%%%%%%%%%%%%%%%%%%%%%%%%%%
\section{Discussion}
We show that Low-Rank Adaptation (LoRA) can systematically map phenotypic variation to modulations of the regulatory parameters of Neural Cellular Automata (NCAs).
Rank-one adaptations suffice to implement horizontal and vertical scaling of a baseline phenotype, amounting to only $\approx 2.6\%$ of the NCA's total parameter count. Their linear modulation and combination bias the nonlinear multicellular dynamics that generate coherent, continuous Cartesian transformations beyond the conditions used for training. 
The same scaling adaptations transfer zero-shot across morphologies: when applied to NCA developmental programs that share the same underlying regulatory network---the scaffold---but grow different phenotypes, they rescale all tested targets without further training while largely preserving their internal features.
D'Arcy Thompson's grid transformations can thus be represented---and controlled---by interventions along low-dimensional directions in the regulatory weight space of a cybernetic developmental material.\cite{thompson1917growth}
Biologically, interventions along such directions model tunable and composable modulations of cellular gene-regulatory and biophysical networks, suggesting how regulatory changes---whether induced by changing environmental conditions or selected over evolutionary time---could coherently redirect organismal form during development.
Howard Pattee proposed that descriptions at higher levels in a self-regulatory hierarchical system arise through a selective loss of microscopic detail, retaining only what is relevant to the corresponding collective function; in turn, constraints at the collective level can act back on lower-level dynamics without determining every microscopic degree of freedom.\cite{pattee1973control}
The low-dimensional regulatory adapters identified here realize this principle computationally: by modifying the shared regulatory rules governing how cells communicate and compute, they change macroscopic features of NCA phenotypes without micromanaging all cellular details or trajectories, thereby providing concrete implementations of such multiscale control handles.

The shared scaffold provides a common regulatory reference frame in which phenotype-specific adaptations of the regulatory machinery---i.e., the LoRA weight modulations of a reference frame NCA---can be compared.
We trained approximately $N_\phi\approx25,000$ phenotype-specific LoRA-NCAs relative to the same frozen scaffold and analyzed their effective weight adaptations, $\Delta\boldsymbol{W}_\phi=\boldsymbol{A}_\phi\boldsymbol{B}_\phi$.
The resulting regulatory weight space exhibits a semantic geometry:
Its first two principal components correlate with horizontal and vertical phenotypic extent, while mean differences between selected subsets yield directions associated with style and symmetrical fission.
These relations are not merely descriptive but functional across morphologies.
Adding the corresponding directions to phenotype-specific adaptations causally changes the NCA dynamics in a controlled and reusable way, compressing or stretching developmental outcomes, transferring stylistic features, or splitting and fusing their morphology.
The shared scaffold thus provides a common language for the fine-tuned cellular interactions through which variations in macroscopic phenotypic features become discoverable and actionable directions in regulatory weight space.

For future bioengineering approaches, such functional regulatory directions provide systems-level interventions into the regulatory machinery of cybernetic material.
As experimenters, we either learn such directions by gradient-based optimization toward a specified anatomical transformation or discover them empirically in a semantically structured weight space. In either case, the resulting distributed but low-dimensional modulation of the shared local interaction rule of the cellular agents biases the collective toward a corresponding anatomical outcome without prescribing individual cellular actions or targeting a single effector.
This suggests a systems-biology strategy for bioengineers, in which desired phenotypic outcomes are addressed through distributed but tailored regulatory modulations, while cells resolve the detailed developmental trajectory through their ongoing regulation and communication.
NCAs provide a minimal model in which this form of control can be studied, but their ANN weights cannot yet be assigned directly to particular genes, morphogens, physiological signals, or developmental events.\cite{Hartl2025NCAs}
Which nodes and interactions in biological gene-regulatory, physiological, and bioelectric networks provide equivalent control handles therefore remains an open question.\cite{Davidson2002,levin2018bioelectriccode}
Identifying such handles in physically and biologically grounded models could provide a principled route toward redirecting developmental attractors in regeneration and developmental repair.\cite{Levin2024MultiscaleWisdom}

The explicitly learned hyper-directions of scale and the corresponding empirically derived PCA directions are only weakly aligned in weight space, with $\lesssim10\%$ cosine similarities, despite both controlling horizontal and vertical phenotypic extent.
This demonstrates degeneracy of morphogenetic control: distinct directions in parameter space can modulate the same macroscopic phenotypic feature.\cite{Edelman2001Degeneracy,conrad1990evogeom,Newman2019}
Moreover, the scale, style and fission directions explored in this work show how several functional organizations can coexist as ``orthogonal'' control handles within the same dynamical substrate, reminiscent of poly-computing.\cite{bongard_2023_plentyofroom}
Together, this can support evolvability by providing multiple regulatory paths toward phenotypic variation while retaining the traits and proportions exposed to little selection pressure.\cite{Wagner1996, Gerhart2007}

These functional regulatory handles and especially their transferability across phenotypic contexts
therefore provide a minimal computational analogue of facilitated variation:\cite{Wagner1996, Gerhart2007}
conserved regulatory dynamics can be retained, while compact modulations expose coordinated changes in system-level growth, form, and function as effective handles to evolutionary search.
If accessible to heritable variation, an organism's reproductive machinery could couple environmental (or physiological) pressures to transmissible modulations of these regulatory handles.
Movement along the corresponding hyper-directions would then provide a compact action space through which offspring development could be biased toward coherent changes in better-fitted organismal form.
Moreover, regulatory handles acquired in one context---and represented by a particular regulatory hyper-direction---would not have to be rediscovered independently in every other context.
They could instead be retained, reused and recombined by evolution's reproductive machinery, providing a combinatorial advantage and thereby improving evolvability:
the range of accessible phenotypic variation would grow through recombination of existing capabilities, rather than only through the acquisition of new ones by point mutations.\cite{Wagner1996,Wagner2007,Gerhart2007}
Stacked LoRA hyper-directions provide a minimal computational setting for testing whether such combinations increase the efficiency and range of phenotypic variation accessible to evolution.
In turn, regulatory capabilities evolved separately in different lineages may be brought together through symbiogenesis and integrated into a novel functioning whole.\cite{barricelli1954esempi, barricelli1954esempiEn, barricelli1957symbiogenetic, Sheldrake2021, aguerayArcas2025whatIsIntelligence, Ashford2026Symbiogenesis}

Repeated environmental or physiological stresses might cause regulatory organization to ``give way'' along directions that reduce those stresses and, if coupled to inheritance, bias offspring development accordingly---a process related to adaptation by natural induction rather than random search.\cite{Buckley2024NaturalInduction, Watson2025NaturalInductionI, Watson2025NaturalInductionII}
Whether biological reproductive machinery operates through such stress-guided facilitated adaptation or can even more actively select and combine regulatory handles according to their functional effects remains an important open question.

We have tackled D'Arcy Thompson's \textit{On Growth and Form}\cite{thompson1917growth} by learning how to navigate the space of \textit{regulatory function} of the cellular agents that collectively generate phenotypic shape transformations in a minimal cybernetic tissue.
The resulting regulatory weight space exposes effective directions for dominant transformations, but both the physical meaning and accurate semantic interpretation of that space remain partly elusive.
Leading principal components capture dominant features shared across many phenotypes, including size, color and overall style, whereas high-fidelity reconstruction of individual NCA dynamics requires thousands of components, as detailed in \Cref{app:lora:pca}.
Moreover, the extracted directions are not fully feature-specific:
the leading scale directions also alter brightness, while the established style-transfer direction simultaneously changes the phenotypic size.
The present weight space is therefore not yet sufficiently organized to isolate detailed, feature-specific regulatory modulations in general, nor can the identified hyper-directions be assigned a clear biophysical meaning: the NCAs' dynamics unfold on an intrinsically abstract 2D grid without the necessary physical constraints shaping differential growth across successive developmental stages.\cite{Rustarazo2026DevelopmentalTiming, HUXLEY1924}
A biophysical interpretation---for example, as coordinated modulations of regulatory nodes controlling morphogen concentrations, relative growth rates, or developmental timing\cite{Briscoe2017,HUXLEY1924,alberch1979size}---requires explicit physical, chemical, and metabolic modelling.
Future work will therefore explore more structured LoRA representations across high-level and detailed features, drawing on recent advances in open-endedness--driven disentangled representation learning,\cite{Kumar2025Fractured}, epiplexity,\cite{finzi2026epiplexity, Zhang2026Epiplexity} and continuous co-evolving search,\cite{hartl2025hades} while embedding the resulting regulatory handles in more biophysically grounded NCAs, neural particle automata, or related mechanical models.\cite{Hartl2025MS, rojas2024EnergyCreatures, Kim2026NPA}
The longer-term goal is to develop intervenable and potentially predictive digital twins of biological systems.

By establishing a causal relation between macroscopic anatomical variation and microscopic regulatory modulation in a multiscale cybernetic system, this work not only renders Thompson's century-old problem tangible but suggests a path toward the inverse problem: inferring which local rule modulations correspond to a desired anatomical transformation.
Much like modern image or video generators are capable of generating desired output from mere text prompts,\cite{Dhariwal2021DMBeatGANs, ho2021classifierfree, brooks2024videoworldsimulators} conditional generative AI may enable generating tailored developmental programs whose distributed dynamics then grow, maintain, or alter anatomical form or function toward a desired outcome.
Future work will thus be dedicated to implementing such an anatomical compiler.\cite{lagasse2023future}
Extending this approach to biophysically grounded models may help identify coordinated regulatory interventions for regeneration and developmental repair, with considerable implications for future bioengineering: \cite{levin2019endogenous,Levin2024MultiscaleWisdom}
tailored interventions for gene-regulatory or biophysical networks could be generated on the fly, e.g., to guide congenitally malformed tissues toward functional organization, redirect wound closure toward limb-regeneration, tackle aging or rejuvenation, or restore tissue-level coordination when cells escape organism-level control in cancer.\cite{levin2021bioelectrical, pai2022hcn2, lopez2023aging}

%%% CONTENT %%%%%%%%%%%%%%%%%%%%%%%%%%%%%%%
%%%%%%%%%%%%%%%%%%%%%%%%%%%%%%%%%%%%%%%%%%%

%%%%%%%%%%%%%%%%%%%%%%%%%%%%%%%%%%%%%%%%%%%
%%% BACK MATTER %%%%%%%%%%%%%%%%%%%%%%%%%%%
\FloatBarrier
\section*{Acknowledgments}
We thank members of the Levin Lab---especially Léo Pio-Lopez and Axel de Baat---for helpful discussions, and Tomika Gotch for help with the manuscript.
We acknowledge support from Astonishing Labs, Inc., and the Templeton World Charity Foundation, Inc. (TWCF0606).
This publication was made possible through the support of Grant 62212 from the John Templeton Foundation. 
The opinions expressed in this publication are those of the author(s) and do not necessarily reflect the views of the John Templeton Foundation. 
M.M. is funded by the Novo Nordisk Foundation Synergy Grant REPROGRAM number
NNF23OC0086722.
ChatGPT (GPT-5.6, OpenAI) was used in targeted, iterative exchanges to assist with coding and debugging tasks and with language refinement; all AI-assisted code was manually reviewed, revised, and tested, and all AI-assisted text was reviewed and edited by the authors, who take full responsibility for the manuscript and computational results.

\section*{Conflict of Interest}
This research was partially funded at Tufts under a Sponsored Research Agreement with Astonishing Labs. M.L. is a co-founder and shareholder of Astonishing Labs. Astonishing Labs has certain rights to inventions associated with this research.

\section*{Author contributions} 
B.H. conceived the operational idea and led the resulting study, developed and implemented the methods, performed the computational experiments and analyses, prepared the figures, and wrote the original manuscript.
M.M. provided methodological feedback and contributed to the analysis and interpretation of the results.
M.B. contributed critical conceptual feedback.
S.R. provided feedback on the study and manuscript.
M.L. posed the motivating conceptual question, provided ongoing biological guidance throughout the project, and helped refine the manuscript.
All authors reviewed and approved the final manuscript.

%%%%%%%%%%%%%%%%%%%%%%%%%%%%%%%%%%%%%%%%%%%
%%% REFERENCES %%%%%%%%%%%%%%%%%%%%%%%%%%%%
\FloatBarrier
\bibliographystyle{vancouver}
\bibliography{references}
%\bibliographystyle{unsrtnat}
%%% REFERENCES %%%%%%%%%%%%%%%%%%%%%%%%%%%%
%%%%%%%%%%%%%%%%%%%%%%%%%%%%%%%%%%%%%%%%%%%

\newpage
\FloatBarrier
\appendix
\section*{Appendix}
\section{Simulation and Training Details}
\label{app:param}

\subsection{NCA Parameters}
\label{app:param:nca}
All simulations used a 2D growing NCA\cite{mordvintsev_growing_2020} implemented in JAX/Flax NNX using CAX v0.2.1.\cite{faldor2025CAX} 
Each cell carried $C=32$ state channels, with the final four interpreted as RGBA and the alpha channel additionally determining cell viability. 
Fixed identity and horizontal and vertical Sobel gradient filters project the cellular neighborhood into $96$ perceptual features, which were passed through two pointwise convolutional layers ($96\rightarrow96\rightarrow32$) with a ReLU activation and a zero-initialized output kernel to generate residual state updates. 
Cell-wise dropout at rate $0.25$ during training and inference implements asynchronous update through the stochastic binary update mask $m_i^t$; 
viability was thresholded at $0.1$ over a $3\times3$ neighborhood before and after each update, with periodic boundary conditions;
also refer to \Cref{sec:methods:nca} for terminology.
LoRA factors modulated both pointwise kernels, contributing $320r$ parameters at rank $r$---$320$ parameters at rank $r=1$ and $5,120$ at rank $r=16$---relative to $12,416$ parameters in the base update network. 
RGBA targets were normalized to $[0,1]$, resized to $37\times37$ pixels, and padded by 15 pixels on each side, producing a $67\times67$ lattice initialized by one living central seed cell. 
For each target, training maintained a pool of $1,024$ developmental states.\cite{mordvintsev_growing_2020} 
At every update, eight states were sampled, the highest-loss state was replaced by a fresh seed, and each state was evolved for $128$ steps; the mean-squared reconstruction loss was evaluated at a randomly selected step $64\leq t<128$. 
Gradients were clipped to unit norm and optimized using AdamW with weight decay $10^{-4}$ and initial learning rate of $1.1618\times10^{-3}$, linear decay to $0.1$ of this value over $6,000$ optimizer updates and kept constant for the remaining $4,000$ epochs. 

\subsection{Shared-Baseline Training}
\label{app:param:baseline:training}
To obtain the shared regulatory scaffold $\boldsymbol W_0$, we jointly optimized its update-network parameters and rank-$16$ target-specific adapters for $120$ \textit{Noto Emoji} targets. 
The targets were distributed among ten parallel workers, with $12$ targets per worker. 
During each outer iteration, every worker shuffled its targets and processed them in three groups of four. 
Within each group, the adapter corresponding to each target was activated in turn, and the resulting four gradients were averaged before updating the shared scaffold and active adapters. 
Each target retained its own developmental-state pool, while four adapter slots were reused by loading and saving the corresponding target-specific LoRA factors. 
Each worker completed $10,000$ outer iterations, corresponding to $30,000$ four-target optimizer updates. 
Every $100$ iterations, its scaffold was merged into the shared checkpoint according to $\theta_{\mathrm{shared}}\leftarrow0.9\theta_{\mathrm{shared}}+0.1\theta_{\mathrm{worker}}$ and then reloaded, with the optimizer reinitialized while retaining the developmental-state pools. 
After training, the target-specific adapters were removed and $W_0$ was retained as the common, target-independent reference frame.

\subsection{Frozen-Baseline Finetuning}
\label{app:param:lora:training}
All subsequent LoRA-fitting runs, including training on targets of other vendors, constituted a single frozen-baseline fine-tuning stage. 
Starting from the stripped scaffold $\boldsymbol W_0$ (\Cref{app:param:baseline:training}), we froze all shared weights and biases, $\theta_0$, and independently initialized one rank-$16$ LoRA factor for every available emoji--vendor pair. 
Emoji identities were taken from the fully qualified sequences in the Unicode Consortium's versioned \texttt{emoji-test.txt} file,\cite{unicodeEmojiTest} with target artwork obtained from \textit{Noto Emoji}, \textit{Twemoji}, \textit{OpenMoji}, \textit{JoyPixels}, \textit{Blobmoji}, \textit{EmojiOne3}, \textit{EmojiTwo}, \textit{TossFace}, and \textit{Fluent3D}. 
Each LoRA factor $\Delta\boldsymbol{W}_\phi$ was activated independently, optimized for $10,000$ updates using the protocol described above, and saved at its lowest observed training loss. 
Jobs were distributed independently across workers; the resulting dataset therefore comprises $N_\phi\approx25,000$ independently fitted LoRA factors $\Delta\boldsymbol{W}_\phi$ sharing the same scaffold $\boldsymbol W_0$.

\subsection{Imposed Mappings of Phenotypic-Transformation and Regulatory Modulation}
To learn the Cartesian scaling directions from \Cref{sec:transform:xy}, we added two rank-one LoRA factors to the frozen NCA $\boldsymbol{W}_{k=1}=\boldsymbol{W}_0+\Delta\boldsymbol{W}_{k=1}$ generating the $37\times 37$ baseline phenotype ${k=1}$.
One adapter is assigned to horizontal $\boldsymbol{\Delta}_x$ and the other to vertical scaling $\boldsymbol{\Delta}_y$. 
Both adapters were trained jointly for $10,000$ iterations across a set of $5\times5=25$ prescribed target transformations. 
During training, these transformations were randomly shuffled, and each target was paired with the corresponding adapter coefficients $(\beta_x,\beta_y)$ specifying its horizontal and vertical deformation; refer to \Cref{sec:transform:xy} for details. 
Only the two rank-one adapters were optimized; the underlying regulatory scaffold $\boldsymbol{W}_{k=1}$ remained fixed. 
Thus, a single pair of adapters was learned across all training transformations rather than fitting a separate adapter to each target.
An independent pool of size $8$ was used per prescribed target.
Learning rate started at $5\times10^{-4}$, decayed linearly to $10\%$ of this value over $3,000$ steps, and remained constant for the remaining $7,000$ epochs.

\subsection{Regeneration Experiments}
\label{app:lora:regeneration}
The NCA solutions used in this contribution have not been trained for regeneration, but for morphogenesis and morphostasis.
In order to identify weight modulations that allow these non-regenerative NCAs to become regenerative, we tried two approaches.

First, we trained a LoRA factor with an additional non-regenerative NCA as scaffold but use a regenerative loss (with adversarial phenotypic ablations during and after development, cf. \cite{mordvintsev_growing_2020}).
However, the training did not turn out to be fruitful: LoRA did not suffice to make the non-regenerative scaffold significantly more regenerative.

Second, we trained a novel LoRA adapter specific to the phenotype $\Delta\boldsymbol M_{k}$ from the common reference scaffold $\boldsymbol W_0$ and immediately used the regenerative training pipeline specified in \cite{mordvintsev_growing_2020} --- in contrast to the non-regenerative loss used to train $\Delta\boldsymbol W_k$.
The trained NCA, $\boldsymbol W_0 + \Delta\boldsymbol M_k$, indeed showed improved regenerative capabilities for phenotype $k$ compared to the non-regenerative counterpart with weights $\boldsymbol W_0 + \Delta \boldsymbol W_k$.
However, the cosine-similarity between $\Delta\boldsymbol M_k$ and $\Delta \boldsymbol W_k$ is very low ($\approx0$), suggesting that $\Delta\boldsymbol M_k$ implements a qualitatively different regulatory machinery compared to $\Delta\boldsymbol W_k$, despite developing the same phenotype $k$.
Efforts to apply the translation vector $\boldsymbol\Delta_{M}=\Delta \boldsymbol M_k-\Delta \boldsymbol W_k$ to a different NCA, i.e., $\boldsymbol W_0 + \Delta \boldsymbol W_{\phi\neq k}+\boldsymbol\Delta_{M}$ did not result in a regenerative NCA for phenotype $\phi$, but turned out largely dysfunctional.
We leave further investigations to future studies.

\section{Metrics}
\subsection{Deep Features as a Perceptual Metric}
\label{app:metrics:perceptual:loss}
\begin{figure}
    \centering
    \includegraphics[width=\linewidth]{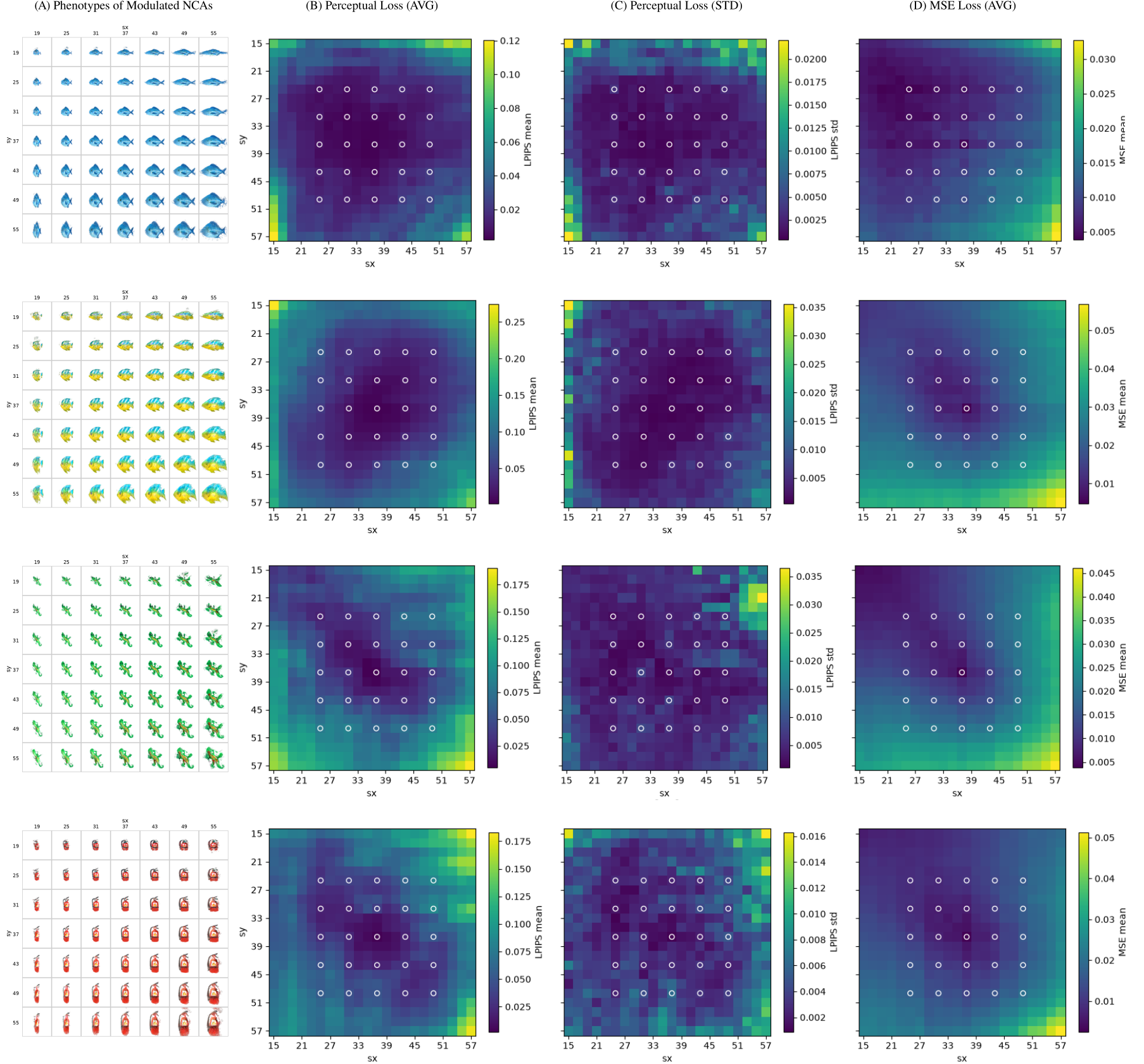}
    \caption{
        \textbf{Low-rank regulatory adaptations induce scale transformations that generalize beyond training conditions and across emoji phenotypes.}
        LoRAs trained on a discrete set of $xy$-scaling transformations for a single phenotype $\boldsymbol{Y}$---with $s_x,s_y\in\{25, 31, 37, 43, 49\}$, see \Cref{fig:lora:transform:xy}---not only generalize zero-shot to out-of-distribution scaling $s_x^\prime,s_y^\prime$, but also transfer across phenotypes, i.e., to NCAs with shared reference scaffold $\boldsymbol{W}_0$ but fine-tuned on a different target $\boldsymbol{Y}_\phi$.
        Column (A) shows developmental outcomes for four different targets across selected sizes $s_x^\prime,s_y^\prime\in\{17, 25, 31, 37, 43, 49, 55\}$---from top to bottom:
        the original fish emoji $k$;
        a similar fish phenotype, but with different color-composition, internal features;
        completely different green lizard phenotype;
        and a red fire extinguisher.
        Column (B) and (C) depict the mean and STD of the perceptual loss~\cite{Zhang2018PerceptualLoss} (LPIPS, see \Cref{app:metrics:perceptual:loss}) per emoji (row) across a fine grid of size transformations $s_x^\prime,s_y^\prime\in[15,57]$; we used $16$ independent stochastic NCA rollouts per (target, $s_x^\prime$,$s_y^\prime$) tuple.
        Column (D) depicts the pixel-wise mean-square error loss between transformed targets and developmental outcomes of the modulated NCAs, which is, however, mostly correlated with the overall phenotypic size rather than actual pixel-wise or semantic accuracy of the modulated developmental outcomes.         
        Especially from the perceptual loss (B) we learn that all of the tested target phenotypes $\phi$ can be scaled via the same hyper-directions $\boldsymbol\Delta_{x,y}$, even beyond training conditions; the white circles in (B-D) emphasize training conditions, where training has only been conducted on the blue fish target.
        This strongly suggests that these LoRAs are universal directions of scale for NCAs with shared scaffold.
        }
    \label{app:fig:lora:generalize:xy}
\end{figure}

Comparing $xy$-transformed phenotypes in \Cref{fig:lora:transform:xy,fig:lora:generalize:xy} via MSE loss mostly correlates with transformation size, $s_x,s_y$, and gives limited information about the actual pixel-wise or semantic quality of the transformed phenotype, see \Cref{app:fig:lora:generalize:xy}~D.
Hence, we use the perceptual loss proposed by \cite{Zhang2018PerceptualLoss} and compare the similarity of encoded target features $z_{\boldsymbol{Y}_k}(s_x, s_y)=E_A(\boldsymbol{Y}_k(s_x, s_y))$ and developmental outcomes $z_{\boldsymbol{X}_k}^t(s_x, s_y)=E_A(\tilde{\boldsymbol{X}}^t_k(s_x, s_y))$ in a semantically rich latent space, using ``alexnet'' as the encoder $E_A$, as proposed by \cite{Zhang2018PerceptualLoss}.
The average perceptual loss, the STD of the perceptual loss, and the average of the MSE loss for out-of-distribution scaling and cross-phenotype experiments can be seen in \Cref{app:fig:lora:generalize:xy}.

\subsection{Principal Component Analysis}
\label{app:lora:pca}
We here investigate the PCA of the LoRA-NCA dataset from \Cref{sec:lora:pca}.
The LoRA decomposition is not unique, but the individual rank components of $\boldsymbol{A}_k$ and $\boldsymbol{B}_k$ are subject to the following gauge symmetry:
Any invertible matrix $\boldsymbol{R}:\boldsymbol{R}\,\boldsymbol{R}^{-1}=\mathbb 1$ leaves the product $\boldsymbol{A}_k\boldsymbol{B}_k=(\boldsymbol{A}_k\boldsymbol{R})(\boldsymbol{R}^{-1}\boldsymbol{B}_k)=\Delta\boldsymbol{W}_k$ unchanged.
The LoRA gauge can be fixed for every NCA layer via singular value decomposition (SVD) of the effective adaptation
\begin{equation}
    \Delta\boldsymbol W_k^{(l)}=\boldsymbol U_k^{(l)}\boldsymbol\Sigma_k^{(l)}\boldsymbol V_k^{(l)\top}
    \qquad
    \bar{\boldsymbol A}_k^{(l)}=\boldsymbol U_k^{(l)}\boldsymbol\Sigma_k^{(l)1/2},
    \qquad
    \bar{\boldsymbol B}_k^{(l)}=\boldsymbol\Sigma_k^{(l)1/2}\boldsymbol V_k^{(l)\top},
\end{equation}
where the singular modes are ordered by decreasing singular value magnitude $|\sigma_i|\geq|\sigma_{i+1}|$ and a fixed sign convention for their directions is used.
The resulting SVD-canonicalized factors $\bar{\boldsymbol A}_k^{(l)}$ and $\bar{\boldsymbol B}_k^{(l)}$ depend only on $\Delta\boldsymbol W_k^{(l)}$, rather than on the original LoRA basis, ensuring $\boldsymbol A_k^{(l)}\boldsymbol B_k^{(l)}=\bar{\boldsymbol A}_k^{(l)}\bar{\boldsymbol B}_k^{(l)}=\Delta\boldsymbol W_k^{(l)}$.

\begin{figure}[h]
    \centering
    \includegraphics[width=\linewidth]{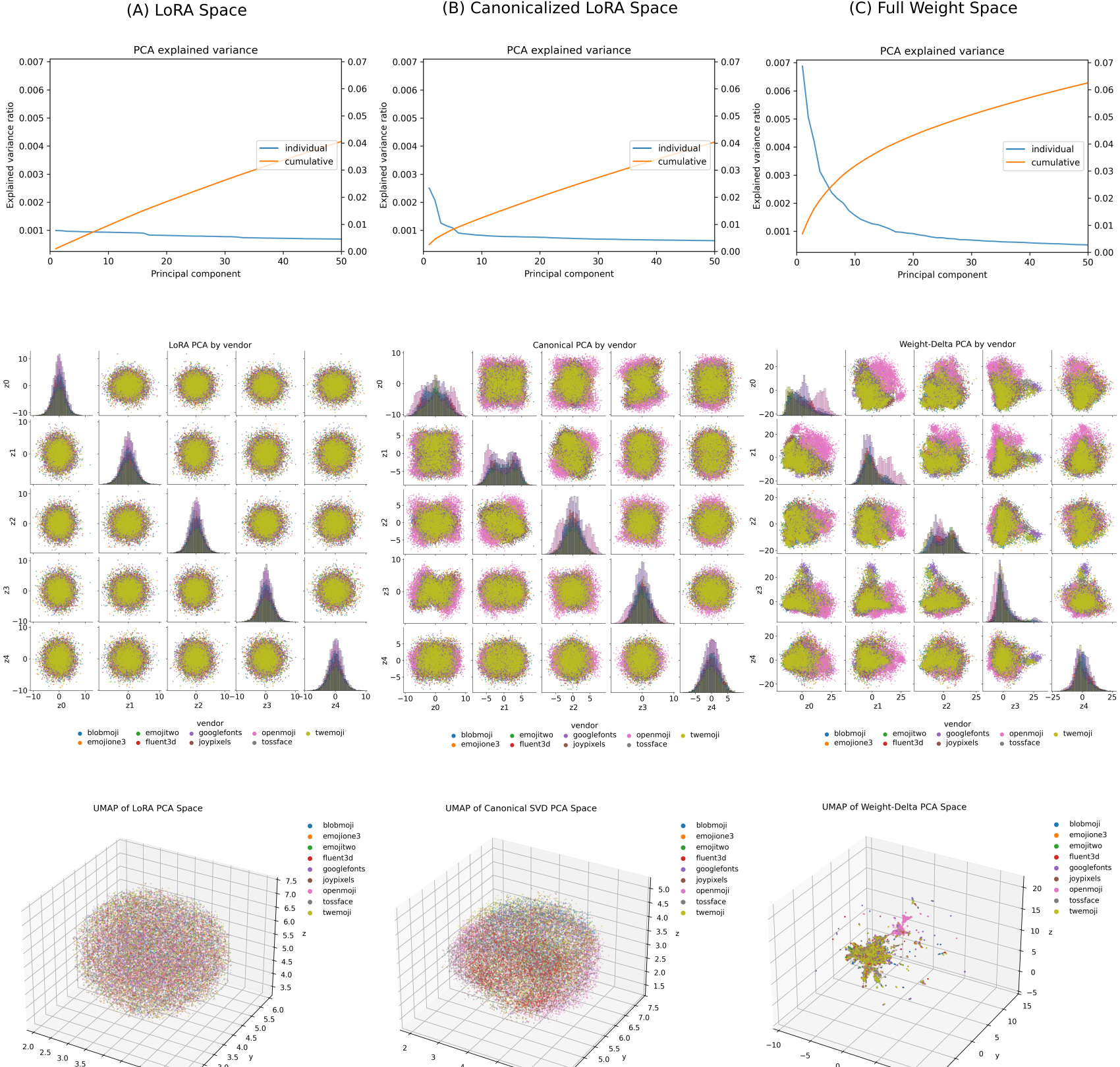}
    \caption{
        \textbf{Principal Component Analysis (PCA) of the LoRA-NCA Dataset from \Cref{sec:lora:pca}.}
        Top/middle/bottom rows, from left to right: Variance explained/PCA-projection/UMAP-projection of (A) raw LoRAs, (B) SVD-canonicalized LoRAs, and (C) full weight-space.
        Prior to PCA, parameters are flattened, concatenated, and standardized to zero-mean and unit-variance per dimension.
        }
    \label{app:fig:lora:nca}
\end{figure}

\Cref{app:fig:lora:nca} compares the PCA of raw LoRA factors, SVD-canonicalized LoRA factors, and full weight-space representations for the LoRA-NCA dataset from \Cref{sec:lora:pca}.
In all cases, we concatenate the flattened set of corresponding parameters and standardize to zero-mean and unit-variance per dimension prior to PCA.
The raw LoRA factors don't separate well when projected onto the leading principal components (PCs).
The SVD-canonicalized LoRAs, and especially the full weight space representation shows considerably more semantic structure.

\begin{figure}[h]
    \centering
    \includegraphics[width=\linewidth]{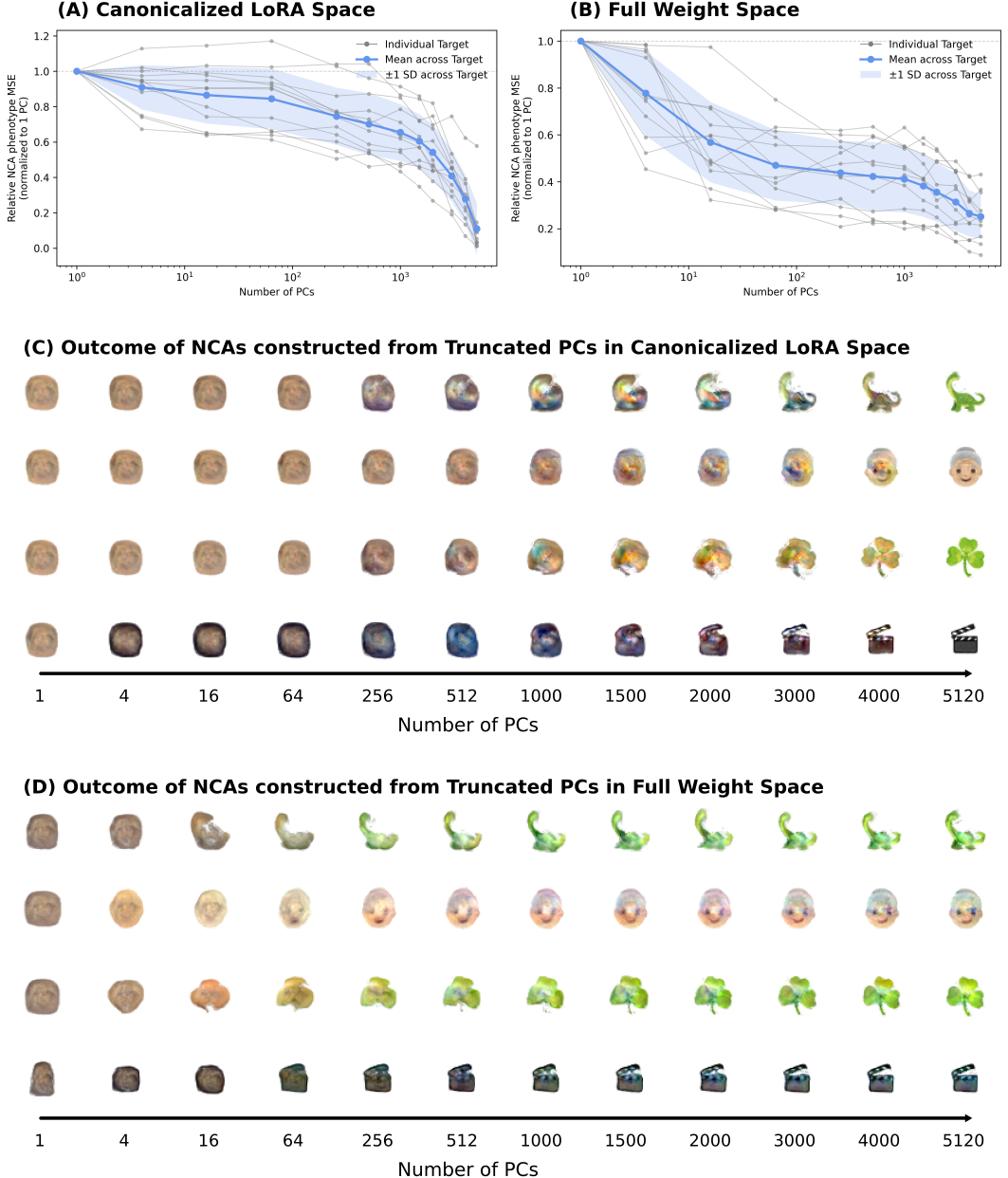}
    \caption{
        \textbf{Outcomes of NCAs constructed with Truncated Numbers of Leading Principal Components (PCs) of the LoRA-NCA Dataset from \Cref{sec:lora:pca}.}
        (A) Mean square error (MSE) of NCA outcomes whose weights are truncated linear combinations of the leading PCs (horizontal axis) of canonicalized LoRA factors (cf. \Cref{app:fig:lora:nca}~B) for individual (grey) and averaged across (blue) multiple randomly chosen target morphologies, respectively normalized to MSE with one leading PC.
        Thousands of PCs are necessary to reach good accuracy; we report similar results for PCs of raw LoRAs, cf. \Cref{app:fig:lora:nca}~A.
        (B) Same as (A) but for the full weight space (cf. \Cref{app:fig:lora:nca}~C).
        Accuracy improves with tens and hundreds of PCs already, but thousands of PCs are necessary to get to high-fidelity reconstructions of the full NCA dynamics.
        (C) Randomly chosen example developmental outcomes (vertical axis) of NCAs constructed from leading PCs (horizontal axis) of canonicalized LoRA factors, cf. (A).
        (D) Same as (C) but for the full weight space, cf. (B).        
        }
    \label{app:fig:nca:truncated}
\end{figure}

We report that high-fidelity reconstruction of the NCAs' dynamics from truncated principal component representations of their parameters requires a considerable number of PCs (\Cref{app:fig:nca:truncated}).
While few leading PCs suffice to cover the rough morphological structure of a given target, the fine details often require thousands of PCs, especially of LoRA or canonicalized LoRA parameters.
The full weight space parameters seem to be more semantically structured, as phenotypic details emerge already from a few hundred PCs, but restoring the high-fidelity outcomes of the full NCA yet again requires thousands of PCs.
This suggests that dominant PCs cover commonly shared dominant features such as size, color, or overall style (cf. \Cref{fig:lora:pca}),
but the semantic relations of detailed features of different phenotypes aren't represented well in the weight space dataset.
We suspect that this is due to the still somewhat ambiguous nature of the shared scaffold:
its main task is to provide a shared reference frame for the phenotype-specific LoRAs,
but we did not regularize whether this reference frame should hold for rough features or detailed developmental pathways.
Future work will resolve how the latter issue can be balanced to provide a semantically structured weight space representation of developmental programs.

\subsection{Twin-Score}
\label{app:lora:pca:twinscore}
Here, we define the twin-score used in \Cref{fig:lora:pca}~E to detect 2D emoji phenotypes consisting of two similarly shaped and similarly sized components.
We first identify all connected components of the foreground mask, treating pixels that touch along an edge or corner as connected, and discard small fragments occupying less than $5\%$ of the original total foreground area.
Samples with other than two remaining components receive a twin-score of $T=0$.
Let $C_1$ and $C_2$ denote the two remaining components and $|C_i|$ their foreground-pixel areas.
Their size balance is given by
\[
b=\frac{\min(|C_1|,|C_2|)}{\max(|C_1|,|C_2|)}.
\]

To compare their shapes, each component is translated into its own local coordinate frame, and its minimal rectangular support is resized to a $32\times32$ binary mask using nearest-neighbor interpolation, yielding $\hat C_i$.
The resulting masks are compared using intersection over union (IoU),
\[
\operatorname{IoU}(A,B)
=\frac{|A\cap B|}{|A\cup B|},
\]
i.e., the fraction of foreground pixels shared by both masks among those belonging to either mask.
Including a left--right mirror reflection, the shape similarity between $C_1$ and $C_2$ is
\[
s=\max\!\left\{
\operatorname{IoU}(\hat C_1,\hat C_2),
\operatorname{IoU}\!\left(\hat C_1,\operatorname{flip}_x(\hat C_2)\right)
\right\},
\]
where $\operatorname{flip}_x$ denotes reflection along the horizontal coordinate.

We define the twin-score as $T=b\,s\in[0,1]$, with high values indicating two equally sized components of similar or mirror-symmetric shape.

%%% BACK MATTER %%%%%%%%%%%%%%%%%%%%%%%%%%%
%%%%%%%%%%%%%%%%%%%%%%%%%%%%%%%%%%%%%%%%%%%
\end{document}